\documentclass{article}\usepackage[T1]{fontenc} 
\usepackage{iclr2025_conference,times}

\usepackage{amsmath,amsfonts,bm}

\newcommand*{\sys}{\textsc{AdvPaint}\xspace}

\def\eqref#1{equation~\ref{#1}}

\def\1{\bm{1}}

\DeclareMathAlphabet{\mathsfit}{\encodingdefault}{\sfdefault}{m}{sl}
\SetMathAlphabet{\mathsfit}{bold}{\encodingdefault}{\sfdefault}{bx}{n}

\usepackage{hyperref}
\usepackage{url}
\usepackage{graphicx}
\usepackage{amsmath}
\usepackage{booktabs}
\usepackage{xspace}
\usepackage{makecell}
\usepackage{amssymb}
\usepackage{mathtools}
\usepackage{wrapfig}

\usepackage{algorithm}
\usepackage{algpseudocode}

\newcommand{\Skip}[1]{}

\title{AdvPaint: Protecting Images from Inpainting Manipulation via Adversarial Attention Disruption}

\author{Joonsung Jeon, Woo Jae Kim, Suhyeon Ha, Sooel Son\thanks{Co-corresponding authors} ~\& Sung-Eui Yoon\footnotemark[1] \\
Korea Advanced Institute of Science and Technology (KAIST) \\
\texttt{\{mikeraph,wkim97,suhyeon.ha,sl.son,sungeui\}@kaist.ac.kr} 
}

\def\eg{\textit{e.g.}}
\def\ie{\textit{i.e.}}

\def\etal{\textit{et al.}}

\iclrfinalcopy 
\begin{document}

\maketitle

\begin{abstract}
The outstanding capability of diffusion models in generating high-quality images poses significant threats when misused by adversaries. In particular, we assume malicious adversaries exploiting diffusion models for inpainting tasks, such as replacing a specific region with a celebrity.
While existing methods for protecting images from manipulation in diffusion-based generative models have primarily focused on image-to-image and text-to-image tasks, the challenge of preventing unauthorized inpainting has been rarely addressed, often resulting in suboptimal protection performance.
To mitigate inpainting abuses, we propose \sys, a novel defensive framework that generates adversarial perturbations that effectively disrupt the adversary's inpainting tasks. \sys targets the self- and cross-attention blocks in a target diffusion inpainting model to distract semantic understanding and prompt interactions during image generation. 
\sys also employs a two-stage perturbation strategy, dividing the perturbation region based on an enlarged bounding box around the object, enhancing robustness across diverse masks of varying shapes and sizes. 
Our experimental results demonstrate that \sys's perturbations are highly effective in disrupting the adversary's inpainting tasks, outperforming existing methods; \sys attains over a 100-point increase in FID and substantial decreases in precision. The code is available at~\href{https://github.com/JoonsungJeon/AdvPaint}{https://github.com/JoonsungJeon/AdvPaint}.

\end{abstract}

\section{Introduction}


The advent of diffusion models~\citep{DDPM, DDIM, LDM} and their applications has enabled the generation of a plethora of highly realistic and superior-quality images. 
For image-to-image tasks~\citep{LDM}, users input an image into a diffusion model to generate a modified version that aligns with a specified prompt.
In inpainting tasks~\citep{LDM}, a diffusion model takes an input image with a masked region and replaces the masked area with new content that reflect a given prompt.

Meanwhile, the technical advancements in diffusion models have also posed significant threats of abuse due to their potential misuse. Unauthorized usage of diffusion models has raised copyright infringement concerns~\citep{news1, news3} and has been exploited to spread fabricated content in fake news distributed across the Internet and social media platforms~\citep{news2}.
To mitigate this abuse, previous research has explored leveraging adversarial perturbations injected into images under protection. These perturbations aim to disrupt subsequent image manipulation tasks involving diffusion models. 

\begin{figure}  
    \begin{center}
    \includegraphics[width=\linewidth, trim={0.1cm 1.5cm 0.2cm 1.3cm}, clip]{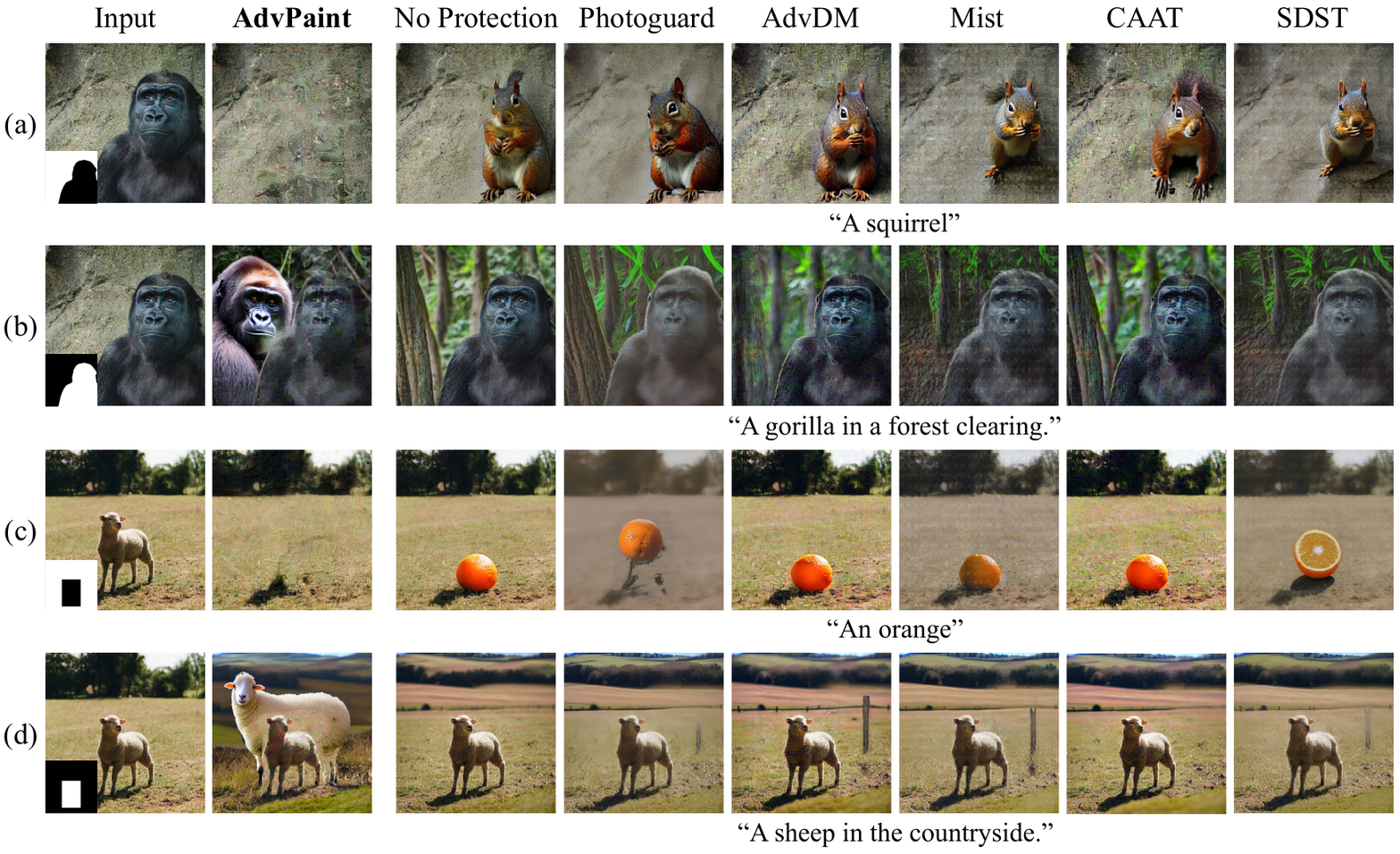}
    \end{center}
    \vspace{-5mm}
    \caption{Our proposed method effectively degrades the result images against various inpainting manipulations with huge spatial differences (\textit{e.g.} removing objects or inserting new objects). The state-of-the-art adversarial examples show limitations in protecting input images, as the generated outputs still harmonize with the prompts. We apply (a) a segmentation mask $m^{seg}$ , (b) its inverse, (c) a bounding box mask $m^{bb}$, (d) and its inverse, respectively.}
   \label{fig:prior works}
   \vspace{-5mm}
\end{figure}

However, the challenge of protecting images from inpainting abuses in diffusion models has received little attention in prior research. We posit that crafting adversarial perturbations to disrupt adversaries' inpainting tasks remains a challenging task.
Figure~\ref{fig:prior works} shows that prior defensive methods of injecting adversarial perturbations~\citep{Photoguard, AdvDM, Mist, CAAT, SDST} provide insufficient protection, allowing adversaries to successfully perform inpainting despite the applied defenses.

We attribute this insufficient protection to the intrinsic nature of inpainting tasks, which leverage a mask for targeted image manipulation. When adversaries conduct foreground inpainting---replacing a \textit{masked} region in an input image with new content specified by a prompt---the perturbations in the \textit{unmasked} area should significantly disrupt this process. 
Similarly, when an inverted mask is applied for background inpainting, the perturbations in the \textit{unmasked} foreground should disrupt the generation process in the \textit{masked} background.

However, prior perturbation methods are designed to protect against whole-image manipulations by applying a single perturbation across the entire image.
This approach is ineffective for inpainting tasks, where only the perturbations in the \textit{unmasked} regions remain, leaving the \textit{masked} areas vulnerable to manipulation by the adversary.
Moreover, previous adversarial objectives have overlooked disrupting the diffusion model's ability to understand the semantics and spatial structure of the image and its conditioning prompts. Instead, they focused on finding shortcuts to make the latent representation of their adversarial examples similar to specific target representations, limiting their effectiveness in inpainting tasks.

In this work, we propose \sys, a novel defensive framework designed to protect images from inpainting tasks using diffusion models. 
\sys crafts imperceptible adversarial perturbations for an image by optimizing the perturbation to disrupt the attention mechanisms of a target inpainting diffusion model. 
Specifically, \sys targets both the cross-attention and self-attention blocks in the inpainting model, ensuring that the perturbation disrupts both foreground and background inpainting tasks. 
Unlike prior works that focused on latent space manipulation, \sys takes a direct approach to the generation process, ensuring that perturbations in the \textit{unmasked} region effectively disrupt the inpainting process in the \textit{masked} region.

The key idea of \sys is to maximize the differences in the components of the self- and cross-attention blocks between a clean image and its corresponding adversarial example, while ensuring the added perturbation remains imperceptible. For cross-attention blocks, \sys perturbs the query input to break the alignment between the latent image representation and external inputs. For self-attention blocks, \sys perturbs the three key elements---query, key, and value---to disrupt the inpainting model’s ability to learn semantic relationships within the input image.

Moreover, \sys divides the perturbation regions into two using an enlarged bounding box around the object in the image and applies distinct perturbations inside and outside the box. This approach ensures that the adversarial examples enhance protection and remain robust to varying mask shapes, making the perturbations more effective and adaptable across diverse inpainting scenarios.

We demonstrate the effectiveness of our approach for both background and foreground inpainting tasks using masks varying in shapes and sizes, achieving superior performance compared to previous adversarial examples. Additionally, we conduct experiments on adversarial examples optimized with three objective functions from prior works, 
and validate the efficacy of our attention mechanism-based approach. 

In summary, our contributions are as follows:
\begin{itemize}

\item We propose the first adversarial attack method designed to disrupt the attention mechanism of inpainting diffusion models, improving the protection of target images against inpainting manipulation abuses by adversaries.

\item We introduce a method for dividing the perturbation region into two, based on the enlarged bounding box around the object in an image under protection, which contributes to \sys remaining effective across diverse adversarial scenarios exploiting various masks and inpainting types.

\item We conduct extensive evaluations on \sys and demonstrate its superior performance compared to other baselines, improving FID by over a 100-point, and building upon existing methods by further enhancing protection specifically against abusive inpainting tasks.
\end{itemize}

\section{Related Studies}
\subsection{Adversarial Attack}

\cite{FGSM} highlighted the vulnerability of neural networks to adversarial examples—inputs perturbed to induce misclassification. 
Kurakin~\etal~extended this by introducing the Basic Iterative Method~\citep{BIM}, which applies small perturbations iteratively, generating stronger adversarial examples. This iterative approach sparked further advancements in attack strategies, leading to more sophisticated methods and defenses.
One notable advancement in this domain is Projected Gradient Descent (PGD) \citep{PGD}, a more robust iterative variant of FGSM. PGD includes a projection step that ensures the perturbed example $x'_i$ remains within a bounded $\eta$-ball around the original input $x$. Specifically, at each iteration $i$, the perturbed input is updated as follows:
\begin{equation}
    x'_{i+1} = \text{Proj}_\eta (x'_i + \alpha \cdot \text{sign}(\nabla_{x'_i} J(\theta, x'_i, y)),
\end{equation}
where $\alpha$ is the step size, $\text{Proj}_\eta$ enforces the constraint within the $\eta$-ball, $\nabla_{x'_i} J(\theta, x'_i, y)$ is the gradient of the loss function $J$, $\theta$ are the model parameters, and $y$ is the ground truth label. This refinement of FGSM via iterative updates and projection has become a representative method for generating adversarial examples.

\subsection{Diffusion-Based Image Generation and Manipulation}

\cite{DDPM} and \cite{DDIM} laid the foundation for modern diffusion models, with significant advancements like the latent diffusion model (LDM) introduced by \cite{LDM}.
LDMs improve computational efficiency by encoding images $x \in \mathbb{R}^{3 \times H \times W}$ into latent vectors $z_0 \in \mathbb{R}^{4 \times h \times w}$ in lower dimension by the encoder $\mathcal{E}$.
This reduction in dimensionality reduces computational cost while maintaining the model's ability to generate high-quality images.

LDMs consist of two processes: a forward process and a sampling process. In the forward process, Gaussian noise $\epsilon$ is incrementally added to the latent $z_0$ across timestep $t$, transforming it into pure Gaussian noise at $t=T$. The forward process results in a Markov Chain and can be expressed as $z_t = \sqrt{\Bar{\alpha}_t}z_0 + \sqrt{1-\Bar{\alpha}_t}\epsilon$ where $\Bar{\alpha}_t=\prod_{i=1}^t \alpha_i$ is pre-scheduled noise level and $\epsilon \sim \mathcal{N}(0, \mathbb{I})$. The reverse  process, or sampling, denoises $z_t$ back to $z'_0$, using a noise prediction model (often a U-Net) trained to predict the added noise at each $t$. The training objective for the noise prediction is to minimize the following:

\begin{equation}
\mathcal{L}_{noise} = \mathbb{E}_{z_0, t, \epsilon \sim \mathcal{N}(0, \mathbb{I})} \left[\|\epsilon - \epsilon_\theta(z_{t+1} ,t)\|^2_2\right],
\label{eq:noise-loss}
\end{equation}

where $\theta$ represents the parameters of the denoiser $\epsilon_\theta$.
During sampling, the denoised latent at each timestep is computed via $z'_{t-1} = (1/\sqrt{\alpha_t}) \left(z'_t - (1 - \alpha_t)/\sqrt{1 - \bar{\alpha}_t} \cdot \epsilon_\theta(z'_t, t)\right) + \sigma_t \epsilon$, where $\sigma_t$ controls the variance of the noise added back at each $t$, ensuring stochasticity. After the sampling process, the final latent $z'_0$ is decoded back into the image space via a decoder $\mathcal{D}$.

In addition to the default LDM, \cite{LDM} introduced an inpainting-specific variant of the U-Net denoiser (see Appendix~\ref{appendix:architecture}). This model takes as input the original image $x$, the mask $m$, and the masked image $x^{m} = x \otimes m$, where $\otimes$ represents element-wise multiplication. The same encoder $\mathcal{E}$ is employed to create latent vectors $z_0$ and $z_0^m$ from both $x$ and $x^{m}$. 
The denoiser then takes as input a concatenation of the latent variable $z_t$, the masked latent $z_0^m$, and the resized mask $m' \in \mathbb{R}^{1 \times h \times w}$ at each $t$. Here, $z_0^m$ and $m'$ are inserted as input for every $t$, only denoising $z_t$ from $t=T$ to $0$. This structure enables effective reconstruction of \textit{masked} regions while preserving coherence with the surrounding \textit{unmasked} areas. 
The loss function for the inpainting task is modified as minimizing the following:

\vspace{-1mm}
\begin{equation}
\mathcal{L}_{noise} = \mathbb{E}_{z_0, z_0^m, m', t, \epsilon \sim \mathcal{N}(0, \mathbb{I})} \left[\|\epsilon - \epsilon_\theta(z_{t+1}, z_0^m, m', t)\|^2_2\right].
\label{eq:inp}
\end{equation}

\subsection{Adversarial Perturbations to Prevent Unauthorized Image Usage}
Malicious actors are certainly able to exploit the capabilities of diffusion models to generate high-quality and authentic-looking images. 
%
To mitigate such abuses, prior studies have explored methods for injecting imperceptible perturbations into images. These perturbations are designed to disrupt the image synthesis process, preventing diffusion models from effectively manipulating these perturbed images, thus protecting them against unauthorized and harmful usage.
Several recent approaches have focused on protecting images from improper manipulation by leveraging \textit{latent} representations in generative models. PhotoGuard \citep{Photoguard} and Glaze \citep{Glaze} are designed to minimize the distance in latent space between the encoder output, $\mathcal{E}(x+\delta)$, and a target latent representation $z_{trg}$.
The objective is to minimize the following:

\vspace{-2mm}
\begin{equation}
\mathcal{L}_{latent} = \|z_{trg} - \mathcal{E}(x + \delta)\|^2_2,
\label{eq:latent-loss}
\end{equation}

where $\delta$ represents the perturbation applied to the image $x$ to ensure its encoded representation shifts towards the target latent vector $z_{trg}$.

Additionally, recent studies have focused on generating adversarial examples by utilizing the predicted noise from LDMs within the latent space. Anti-Dreambooth~\citep{Anti-Dreambooth} and AdvDM~\citep{AdvDM} adopt $\max_{\delta}$ Equation \ref{eq:noise-loss} to target text-to-image models. MetaCloak~\citep{MetaCloak} introduces a meta-learning framework alongside $\max_{\delta}$ Equation~\ref{eq:noise-loss} to address suboptimal optimization and vulnerability to data transformations. Mist~\citep{Mist} combines the two objectives -- $\max_{\delta}$ Equation~\ref{eq:noise-loss} and $\min_{\delta}$ Equation~\ref{eq:latent-loss} -- to strengthen image protection. CAAT~\citep{CAAT} utilizes $\max_{\delta}$ Equation~\ref{eq:noise-loss} for the optimization of its perturbation, and also finetunes the weights of key and value in the cross-attention blocks. \cite{SDST} propose SDST that enhances the efficiency of optimization by incorporating score distillation sampling (SDS)~\citep{SDS}, which is applied alongside minimizing Equation~\ref{eq:latent-loss}.




\section{Problem Statement}
\label{sec:motive}
\noindent\textbf{Threat model.}
We assume a malicious adversary who attempts to manipulate a published image by conducting inpainting tasks. The adversary’s goal is to perform foreground or background inpainting using publicly available LDMs. They are motivated to fabricate contents using published images to spread fake news, potentially involving important public figures, or to infringe on the intellectual property of published artistic images.

We tackle a research question of how to compose an adversarial example
for a given input image that effectively disrupts inpainting tasks abused by the adversary.
Previous studies~\citep{Photoguard, AdvDM, Mist, CAAT, SDST} have
explored diverse ways of crafting adversarial perturbations that
disrupt adversaries' image-to-image and text-to-image tasks. However,
we argue that these adversarial attacks are insufficient for
protecting images from inpainting tasks.

Inpainting tasks inherently involve leveraging a mask for a specific
target region to replace the \textit{masked} area with desired prompts. However, in this process, the applied mask removes the perturbations embedded in the protected image,
rendering them ineffective in disrupting the adversary's ability to inpaint the protected area.
For example, the inpainted ``orange'' in AdvDM (Figure~\ref{fig:prior works})
exhibits a noisy background in the \textit{unmasked} regions, while
the perturbation has no impact on the generated ``orange" object
itself. We observed a similar limitation in Figure~\ref{fig:prior works} ``for
every prior adversarial works", where the embedded perturbation only
undermines inpainting tasks for the unmasked regions in protected
images. This limitation arises because prior methods focus on optimizing objective functions (Equation~\ref{eq:noise-loss} and \ref{eq:latent-loss}) that seek shortcuts to align the latent representation of adversarial examples with their target representations.

Thus, this limitation presents a technical challenge: adversarial
perturbations in the \textit{unmasked} background should undermine the
generation process in the \textit{masked} area, thereby disrupting
foreground inpainting. The same challenge applies to background inpainting with an inverted mask, where perturbations only on the \textit{unmasked} foreground should disrupt the generation of the background.

To overcome this challenge, we propose two novel methods for
generating adversarial perturbations: (1) generating perturbations that disrupt 
the attention mechanism of inpainting LDMs, and (2) applying distinct perturbations
for a region covering a target object and the surrounding backgrounds outside that region.

These approaches work in tandem to effectively disrupt both foreground
and background inpainting tasks. Section~\ref{sec:att_loss} describes
our adversarial objectives that simultaneously disrupt cross-attention
and self-attention mechanisms of inpainting LDMs. 
In Section~\ref{ss:two-stage}, we describe the optimization strategy, which applies distinct
perturbations to regions inside and outside the bounding boxes encompassing target objects, ensuring robustness across various mask
shapes.

%
%


\section{Methodology}
\label{sec:method}

\subsection{Adversarial Attack on Attention Blocks}
\label{sec:att_loss}

We propose~\sys, a novel defensive framework to protect images from unauthorized inpainting tasks using LDMs.
\sys generates an adversarial perturbation specifically designed to
fool LDM-based inpainting model by disrupting their ability to
capture correct attentions. These perturbations are designed to fundamentally destroy the model's image generation capability by tampering the cross- and self-attention mechanisms simultaneously.

\begin{wrapfigure}[15]{r}{7cm}
\vspace{2mm}
    \includegraphics[width=\linewidth, trim={0.3cm 13cm 16cm 2cm}]{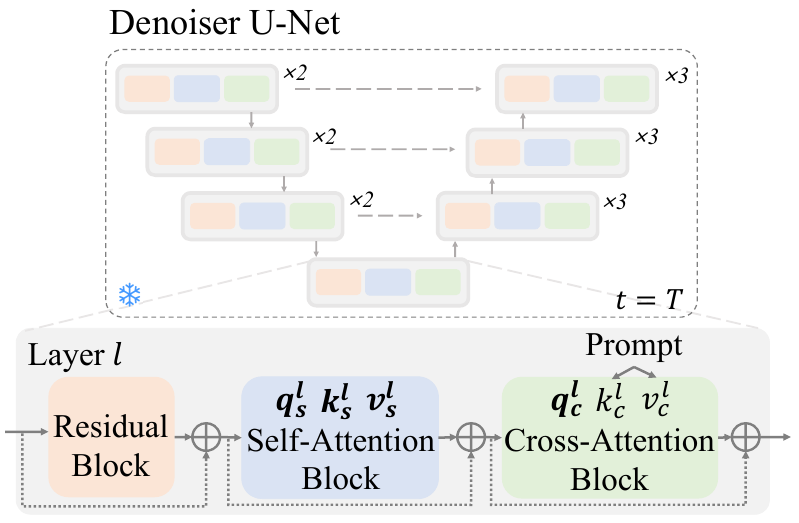}
   \caption{Attention mechanism in the U-Net denoiser. The bolded components in both blocks represent our target for disruption.}
   \label{fig:framework}
\end{wrapfigure}

The attention mechanism in the U-net denoiser $\epsilon_\theta$ of an
LDM reconstructs an image from the latent vector via a denoising
process.
As depicted in Figure~\ref{fig:framework}, each attention block--a
sequence of residual, self-attention, and cross-attention blocks--is
repeated after each downsampling and upsampling operation. The
self-attention block takes three components: query $q$, key $k$, and
value $v$, each of which is a linear transformation of an input image.
The cross-attention blocks obtains $k$ and $v$ sourced from external
conditioning inputs (\eg~prompts). For each layer $l$, $q$
and $k$ are then multiplied to form the attention map $\mathcal{M}$,
which then weights $v$, producing the block output $\mathcal{A}$:
\begin{equation}
\mathcal{M} = \text{Softmax}\left(\frac{qk^T}{\sqrt{d}}\right), \quad \mathcal{A} = \mathcal{M} \cdot v.
\end{equation}
Here, $d$ is the dimension of $q$ and $k$. Against such inpainting LDMs, we propose a novel optimization method for 
crafting adversarial perturbations that disrupt the functionalities of both 
self- and cross-attention blocks.
To attack cross-attention blocks, we design an objective function that
disturbs the alignment between prompt tokens and their spatial
positions in the input image. We aim to maximize the difference
between the query $q$ of the clean image $x$ and that of the
adversarial example $x+\delta$, thereby disrupting the cross-attention
mechanism. Given $q = Q(\phi(x))$, where $\phi(\cdot)$ indicates the
extracted features from the previous layer immediately before the
self- or cross-attention block and $Q(\cdot)$ is the linear projection
operator for $q$, we define the adversarial objective
$\mathcal{L}_{cross}$ as follows:
\begin{equation}
\mathcal{L}_{cross} = \sum\limits_{l} \left\| Q_c^l(\phi(x+\delta)) - Q_c^l(\phi(x)) \right\|^2.
\label{eq:cross-loss}
\end{equation}
Here, $l$ denotes the $l$-th layer in the denoiser, and $c$ refers the cross-attention block. By pushing the query $q$ of $x+\delta$ away from that of $x$,
this objective interferes the alignment with the key $k$ and value
$v$, both of which are derived from the prompt conditions.

We propose another adversarial objective function that specifically
targets the self-attention blocks. Unlike $\mathcal{L}_{cross}$, which
only attacks $q$ that interacts with external conditioning inputs,
this objective targets all input components---$q$, $k$, and $v$---in
the self-attention block. This objective function $\mathcal{L}_{self}$ in
Equation~\ref{eq:self-loss} is designed to maximize the difference
between these three components of $x$ and $x + \delta$:
\begin{equation}
\begin{aligned}
\mathcal{L}_{self} = \sum\limits_{l} \bigg(&  \left\| Q_s^l(\phi(x+\delta)) - Q_s^l(\phi(x)) \right\|^2  \\ 
& + \left\| K_s^l(\phi(x+\delta)) - K_s^l(\phi(x)) \right\|^2  + \left\| V_s^l(\phi(x+\delta)) - V_s^l(\phi(x)) \right\|^2 \bigg).
\end{aligned}
\label{eq:self-loss}
\end{equation}
Here, $K(\cdot)$ and $V(\cdot)$ are linear projectors for $k$ and $v$,
respectively, in the self-attention block $s$, where $k = K(\phi(x))$ and $v =V(\phi(x))$.
By maximizing the difference across all components, we aim to disrupt
the model's ability to interpret the semantics and spatial structure of
the given image.

\subsection{Separate Perturbations for Masked and Unmasked Regions}
\label{ss:two-stage}
We also suggest an additional defensive measures of applying separate perturbations for possible objects that the adversary targets and their surrounding backgrounds.
Specifically, given a target image to protect, \sys first divides the
image into two regions---foreground and background--and applies distinct perturbations to ensure robust protection against masks of varying sizes and shapes.
\sys identifies target objects that the adversary may target using
Grounded SAM~\citep{GroundedSAM}. To achieve this, \sys leverages a
bounding box $m^{bb}$ around the identified objects. To fully cover the objects, this bounding box
is then expanded by a factor of $\rho$ to form a new mask, $m$, by
increasing its height and width while keeping the center coordinates
of $m^{bb}$ intact. 
Using the two masks, $m$ and $1 - m$, \sys computes two separate perturbations for the regions inside and  outside the boundary of $m$ based on the adversarial objective in Equation~\ref{eq:attn-loss}.
\vspace{-1mm}
\begin{equation}
\delta \coloneqq \arg\max_{\|\delta\|_\infty \leq \eta} \mathcal{L}_{attn} =  \arg\max_{\|\delta\|_\infty \leq \eta} \bigg(\mathcal{L}_{cross} + \mathcal{L}_{self}\bigg).
\label{eq:attn-loss}
\end{equation}

In contrast to the single-stage perturbation approach used in prior works, where a single perturbation is applied to the entire image, our two-stage strategy provides enhanced protection and robustness against diverse mask configurations, as demonstrated in Section~\ref{sec:exp_stage} and ~\ref{sec:exp_mask}.

\section{Experiments}

\subsection{Experimental Setup}
\label{Experimental Setup}

We evaluate our proposed method on the pre-trained inpainting model from Stable Diffusion~\citep{LDM}, referred to as \textit{SD inpainter} in this experiment. This model is widely used both in academia (\eg ~\cite{inpaintanything, SDST}) and by the public\footnote{https://huggingface.co/runwayml/stable-diffusion-inpainting}(\eg~\cite{diffusers}).
Following prior studies~\citep{Photoguard, AdvDM, SDST}, we collected 100 images from publicly available sources\footnote{https://www.pexels.com/}\footnote{https://unsplash.com/}, which were then cropped and resized to 512$\times$ 512 resolutions. We applied Grounded SAM~\citep{GroundedSAM} to generate masks of various shapes and sizes. In computing adversarial perturbations, we enlarged the generated bounding box $m^{bb}$ to $m$ by a factor of $\rho = 1.2$, separating the regions for two-stage optimization. 
For text conditions, we generated 50 random prompts using ChatGPT~\citep{ChatGPT}. For example, $\{\text{noun}\}$ and $\{\text{noun}, \text{preposition}, \text{location}\}$ are randomly generated for foreground and background tasks, respectively. 
We applied Projected Gradient Descent (PGD) to optimize our perturbations exclusively at timestep $T$, over 250 iterations, starting with an initial step size of 0.03, which progressively decreased at each step. Importantly, we set $\eta$ as 0.06 for all adversarial examples, including those from prior works, to enforce consistent levels of imperceptible perturbations. All experiments were conducted using a single NVIDIA GeForce RTX 3090 GPU. Further implementation details can be found in the Appendix~\ref{appendix:implementaion details}.

%


\begin{figure}  
    \begin{center}
    \includegraphics[width=0.95\linewidth, trim={0.2cm 4.0cm 2.5cm 1.0cm}, clip]{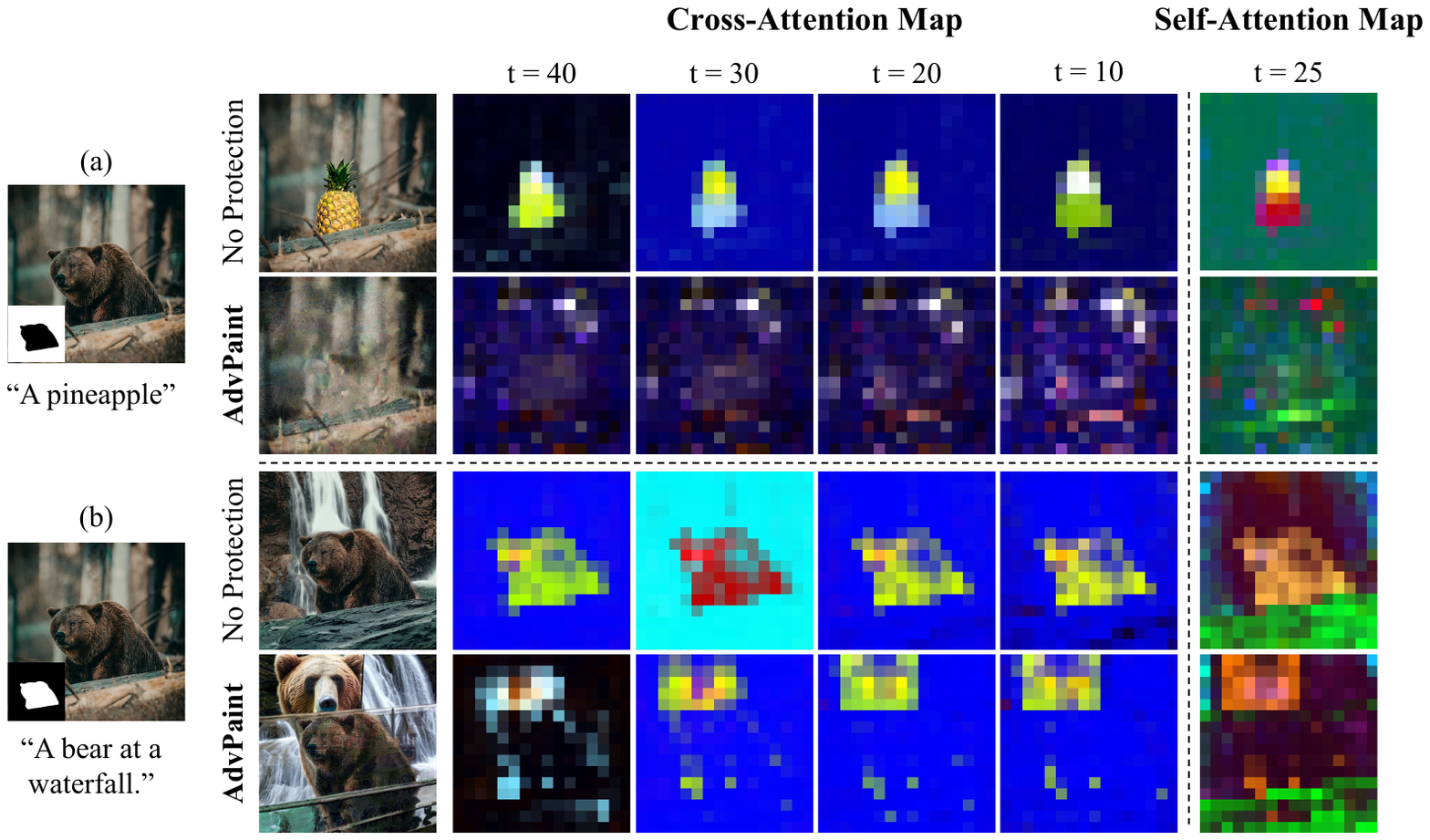}
    \end{center}
    \vspace{-3mm}
    \caption{Visualization of cross- and self- attention maps for (a) foreground and (b) background inpainting manipulations. Our proposed method redirects the model's attention to other regions of the image, as shown in (a), while focusing attention on the newly generated object in (b).}
   \label{fig:attmaps}
   \vspace{-5mm}
\end{figure}

\vspace{-1.5mm}
\subsection{Attention Maps}
\label{sec:exp_att}

Figure~\ref{fig:attmaps} visualizes the cross- and self-attention maps for both the unprotected image $x$ and the adversarial example $x + \delta$ during the inpainting process.
We compare foreground and background inpainting tasks under different textual conditions with the segmentation mask $m^{seg}$.
For the prompts used to produce cross-attention maps, we used ``pineapple" and ``bear", respectively.
The attention maps are visualized using PCA at a specific timestep $t$ within $T = 50$ inference steps. 

In (a) foreground inpainting, the cross-attention map consistently focuses on the ``pineapple'' during the generation process of $x$, but our perturbation redirects attention to other areas of the unmasked region. Our method prevents the model from generating ``pineapple'', producing \textit{nothing in the masked region}.
In the self-attention map, the model shows a scattered image structure and semantics, indicating successful distraction.
On the other hand, in (b) background inpainting, our perturbation causes the model to generate a new ``bear'' in the masked region, disregarding the original ``bear'' in the image. As shown in the cross-attention map, the prompt tokens fail to recognize the original object, leading to \textit{the generation of a new ``bear''}. Additionally, the self-attention block is tricked into overlooking the existing ``bear'', focusing instead on the newly generated one. These distorted attention maps demonstrate the effectiveness of our objective $\mathcal{L}_{attn}$ by (1) disrupting the linkage between image features and prompt tokens in the cross-attention blocks and (2) impairing the semantic understanding in the self-attention blocks.

\vspace{-1mm}
\subsection{Comparison with Existing Methods on Inpainting Tasks}
\label{sec:exp_inpaint}

\begin{figure}
    \begin{center}
    \includegraphics[width=0.95\linewidth, trim={3.2cm 1.3cm 3.2cm 1.3cm}, clip]{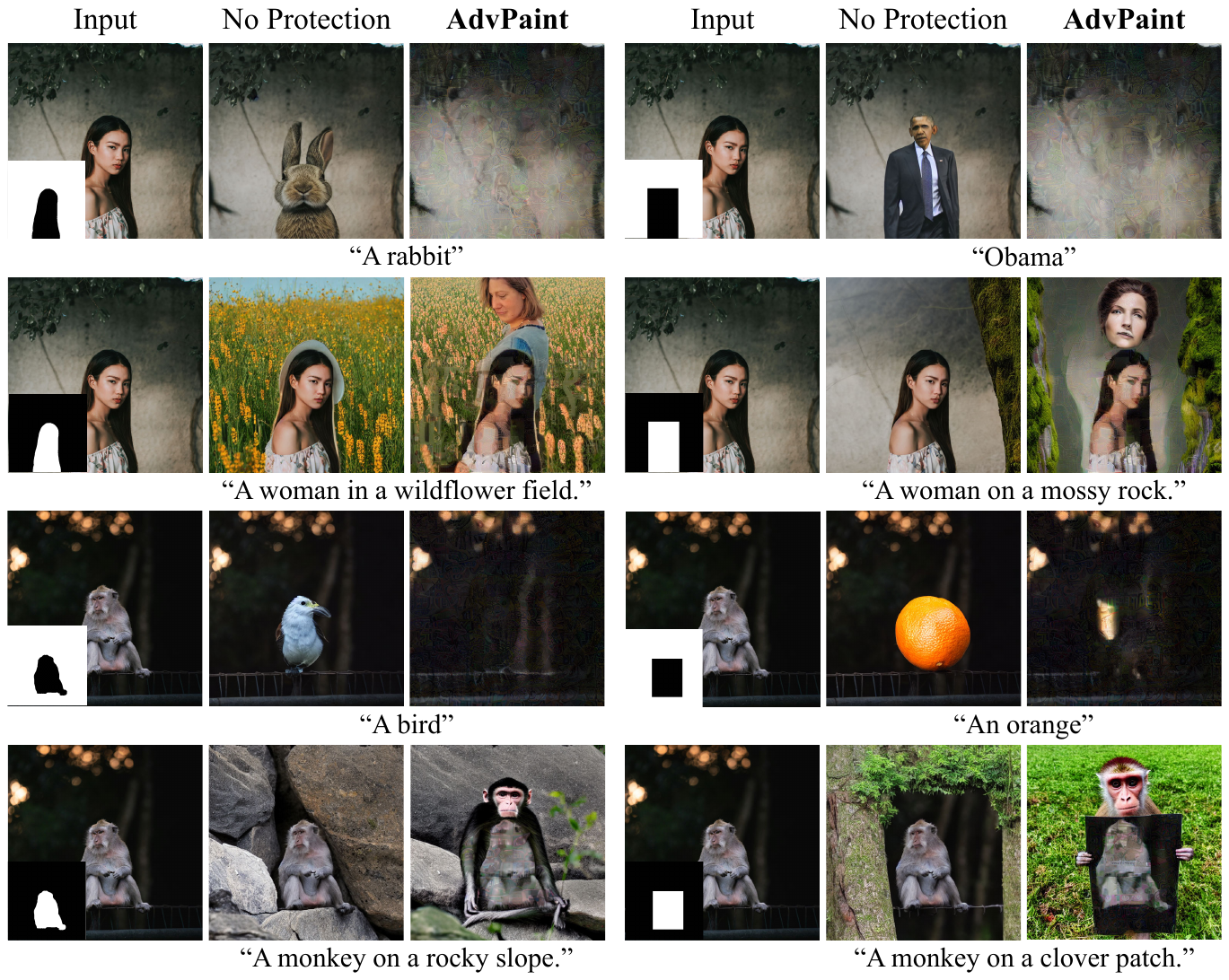}
    \end{center}
    \vspace{-3mm}
    \caption{Our proposed adversarial examples on diverse inpainting types, masks, and prompts.}
   \label{fig:ours}
\end{figure}

We evaluate the performance of state-of-the-art adversarial attack methods, originally designed for disrupting image-to-image and text-to-image tasks~\citep{Photoguard, AdvDM, Mist, CAAT, SDST}, on inpainting tasks.
Qualitative results on inpainting manipulations using the segmentation mask $m^{seg}$ and the bounding box mask $m^{bb}$ are shown in Figure~\ref{fig:prior works}.
For foreground inpainting ((a), (c)), \sys-generated adversarial examples block the creation of objects specified by the given prompts. In contrast, previous methods allow the inpainter model to generate the synthetic objects as described in the prompts. For background inpainting ((b), (d)),  our perturbations successfully mislead the model to ignore the original object and generate a new one from the prompt, while previous methods produce high-quality backgrounds aligned with the prompt. This highlights the challenge for prior approaches, which are less effective at disrupting the generation process within the \textit{masked} area. More qualitative results of \sys are represented in Figure~\ref{fig:ours} and Appendix~\ref{appendix:qual}.

\begin{table}
\huge
\centering
\resizebox{\textwidth}{!}{
\begin{tabular}{l||ccc|ccc|ccc|ccc}
\toprule
 & \multicolumn{6}{c|}{\textbf{Foreground Inpainting}} & \multicolumn{6}{c}{\textbf{Background Inpainting}} \\
\cmidrule(lr){2-7} \cmidrule(lr){8-13}
 & \multicolumn{3}{c|}{$\boldsymbol{m^{seg}}$} & \multicolumn{3}{c|}{$\boldsymbol{m^{bb}}$} & \multicolumn{3}{c|}{$\boldsymbol{m^{seg}}$} & \multicolumn{3}{c}{$\boldsymbol{m^{bb}}$} \\
 
Optimization Methods & FID $\uparrow$  & Prec $\downarrow$ & LPIPS $\uparrow$   & FID $\uparrow$   & Prec $\downarrow$    & LPIPS $\uparrow$    & FID $\uparrow$  & Prec $\downarrow$ & LPIPS $\uparrow$   & FID $\uparrow$   & Prec $\downarrow$   & LPIPS $\uparrow$   \\
\midrule
Photoguard& 230.49 & 0.5244 & 0.6494 & 185.86 & 0.7212 & 0.6236 & 118.85 & 0.4332 & 0.4141 & 132.51 & 0.1844 & 0.5220 \\
AdvDM& 232.39 & 0.3030 & 0.5287 & 181.13 & 0.4794 & 0.5231 &  94.49 & 0.5772 & 0.3111 & 116.60 & 0.2420 & 0.4191 \\
Mist& 235.81 & 0.4590 & 0.5541 & 191.00 & 0.6490 & 0.5421 & 123.48 & 0.4004 & 0.3852 & 155.57 & 0.1602 & 0.5016 \\
CAAT& 232.83 & 0.3430 & 0.5274 & 181.21 & 0.5314 & 0.5192 &  98.22 & 0.5414 & 0.3199 & 118.68 & 0.2382 & 0.4182 \\
SDST& 212.90 & 0.5658 & 0.5042 & 174.85 & 0.7244 & 0.4994 & 112.17 & 0.4406 & 0.3841 & 133.15 & 0.2054 & 0.4809 \\
\midrule
SD Inpainter + $\min_{\delta}$ Eq.~\ref{eq:latent-loss} & 211.35 & 0.5644 & 0.5780 & 180.40 & 0.7214 & 0.5894 & 128.01 & 0.4006 & 0.4745 & 146.39 & 0.1374 & 0.5914 \\
SD Inpainter + $\max_{\delta}$ Eq.~\ref{eq:inp} & 224.81 & 0.3860 & 0.4705 & 199.37 & 0.5186 & 0.4878 & 116.60 & 0.4832 & 0.3844 & 142.37 & 0.2078 & 0.4795 \\
SD Inpainter + $\min_{\delta}$ Eq.~\ref{eq:inp} & 182.12 & 0.6124 & 0.5267 & 154.27 & 0.7560 & 0.5273 &  97.44 & 0.5852 & 0.386 & 107.43 & 0.2692 & 0.4902 \\
\midrule
\sys                     & \textbf{347.88} & \textbf{0.0570} & \textbf{0.6731} & \textbf{289.63} & \textbf{0.1536} & \textbf{0.6762} & \textbf{219.07} & \textbf{0.2148} & \textbf{0.5064} & \textbf{303.90} & \textbf{0.0936} & \textbf{0.6105} \\
\bottomrule
\end{tabular}
}
\caption{Quantitative comparison with existing methods and objectives on foreground and background inpainting tasks. Metrics include FID, Precision (Prec), and LPIPS for segmentation ($m^{seg}$) and bounding box ($m^{bb}$) masks.
}
\label{table:inpainting_comparison}
\vspace{-5mm}
\end{table}

To quantitatively compare our method with prior adversarial examples, we assess the performance using Frechet Inception Distance (FID)~\citep{FID}, Precision~\citep{precision}, and LPIPS~\citep{lpips}. FID and LPIPS measure the feature distance between input images and the generated images. Precision denotes the proportion of generated images that fall within the distribution of real images. We used AlexNet~\citep{alexnet} to compute LPIPS. In adversarial example methods, high FID, low precision, and high LPIPS are preferred.

In Table~\ref{table:inpainting_comparison}, we report the FID, Precision, and LPIPS scores for inpainting tasks using $m^{seg}$ and $m^{bb}$ masks. 
\sys consistently outperforms the state-of-the-art methods across various mask types while maintaining the same level of perturbation budget across all adversarial examples. For instance, when assuming the adversary conducting background inpainting,
\sys using $m^{bb}$ achieves an FID of 303.90, outperforming Mist, the second-best method, by 148.33.
%
We attribute this improvement to the fundamental difference in adversarial objectives. \sys targets the attention blocks and disrupts the functionalities of the self-attention block (image features-to-features) and the cross-attention block (image features-to-prompt) even with partially cropped perturbations. In contrast, the previous works focus solely on latent space representations (as seen in Equation~\ref{eq:noise-loss},~\ref{eq:inp} and~\ref{eq:latent-loss}), without directly influencing the denoiser U-Net during the generation process.

\subsection{Comparison with Prior Objectives for Inpainting Task}
\label{sec:exp_objective}

We assess the effectiveness of adversarial objective functions from prior works on inpainting manipulations by replacing their default LDM (image-to-image or text-to-image) with inpainting-specialized LDM (\textit{SD inpainter}).

Rows 7--9 in Table~\ref{table:inpainting_comparison} show the experimental results for the modified versions of \sys, utilizing the adversarial objectives from prior work (\ie~Equation~\ref{eq:noise-loss} and~\ref{eq:latent-loss}). These results indicate that prior objectives do not lead to significant improvements, even when applied with the same SD inpainter, yielding subpar performance across all metrics. This is likely because previous spatial objectives, which rely primarily on latent space representations, lose effectiveness when image perturbations are masked out before being processed by the inpainting model.




\subsection{Effectiveness of Separate Perturbations for Image Protection}
\vspace{-1mm}
\label{sec:exp_stage}
We conduct quantitative evaluation to assess the effectiveness of \sys using separate perturbations based on the enlarged mask $m$, comparing it with the single perturbation approach.
For the single-perturbation method, a white mask that covers \textit{nothing} in the image is applied. Then, we optimize the single perturbation with the proposed objective $\mathcal{L}_{attn}$. 
Table~\ref{table:stages} compares the protection effectiveness of this single-perturbation method with our approach, which utilizes two masks, $m$ and $1-m$. The results demonstrate that using separate perturbations in \sys provides stronger image protection, yielding significant performance improvements across most metrics, regardless of the inpainting task type.

\begin{table}
\huge
\centering
\resizebox{\textwidth}{!}{
\begin{tabular}{c||ccc|ccc|ccc|ccc}
\toprule
 & \multicolumn{6}{c|}{\textbf{Foreground Inpainting}} & \multicolumn{6}{c}{\textbf{Background Inpainting}} \\
\cmidrule(lr){2-7} \cmidrule(lr){8-13}
 & \multicolumn{3}{c|}{$\boldsymbol{m^{seg}}$} & \multicolumn{3}{c|}{$\boldsymbol{m^{bb}}$} & \multicolumn{3}{c|}{$\boldsymbol{m^{seg}}$} & \multicolumn{3}{c}{$\boldsymbol{m^{bb}}$} \\
 Stage & FID $\uparrow$ & Prec $\downarrow$   & LPIPS $\uparrow$  & FID $\uparrow$  & Prec $\downarrow$  & LPIPS $\uparrow$  & FID $\uparrow$  & Prec $\downarrow$   & LPIPS $\uparrow$  & FID $\uparrow$  & Prec $\downarrow$  & LPIPS $\uparrow$  \\
\midrule
 1 & 345.76 & 0.0628 & \textbf{0.6940} & 271.73 & 0.2056 & \textbf{0.6767} & 191.15 & 0.2418 & 0.4747 & 266.00 & 0.0938 & 0.5936 \\
 2 & \textbf{347.88} & \textbf{0.0570} & 0.6731 & \textbf{289.63} & \textbf{0.1536} & 0.6762  & \textbf{219.07} & \textbf{0.2148} & \textbf{0.5064} & \textbf{303.90} & \textbf{0.0936} & \textbf{0.6105} \\
\bottomrule
\end{tabular}
}
\caption{Performance comparison according to optimization strategy. Combining our proposed objective with two-stage optimization consistently outperforms single-stage optimization.}
\label{table:stages}
\vspace{-2mm}
\end{table}


\begin{figure}  
    \begin{center}
    \includegraphics[width=0.95\linewidth, trim={2.8cm 5.5cm 2.2cm 5.6cm}, clip]{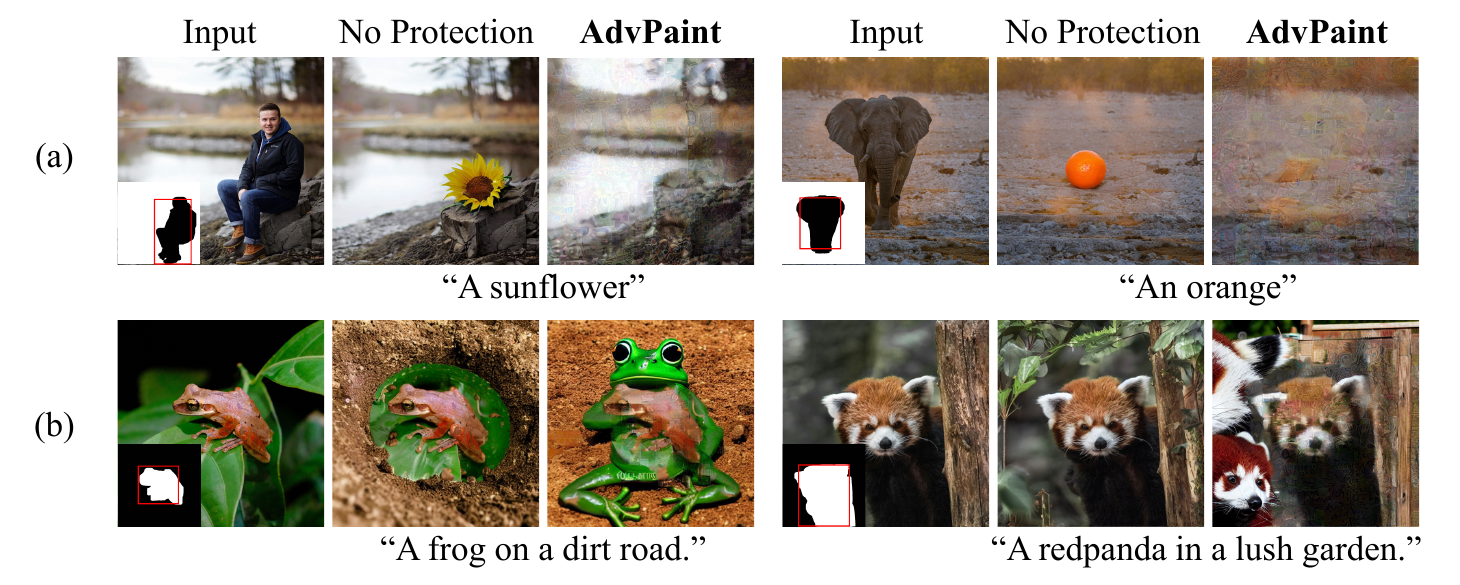}
    \end{center}
    \vspace{-3mm}
    \caption{Results on image inpainting protection with exceeding masks. Hand-crafted binary masks depict real-world scenarios, with red bounding boxes indicating our optimization masks $m$. Examples of both (a) foreground and (b) background inpainting tasks are shown.}
   \label{fig:masks}
   \vspace{-5mm}
\end{figure}

\subsection{Robustness of \sys in Real-World Scenario}
\label{sec:exp_mask}




Although \sys outperforms single perturbation methods, one might question whether \sys remains effective when applied inpainting masks exceed the boundary. Thus, we further evaluate the robustness of \sys in real-world scenarios where masks are hand-crafted and exceed the boundaries of our optimization mask $m$.
To simulate diverse user-defined masks, we randomly shift the original mask and consider two inpainting cases: one where the inference mask extends beyond 
$m$ and another where it remains within $m$. For randomly selected 25 images, we generate 10 segmentation masks per image and randomly shift them up, down, left, or right by a random number of pixels, ensuring that at least one side exceeds the boundary, defining these as $m^{out}$. $m^{in}$ denotes masks that remain within the boundary after the same shifting process. Inpainting manipulations are then performed using 25 prompts for both foreground and background tasks.

\begin{wraptable}{l}{7cm}
\huge
\centering
\resizebox{\linewidth}{!}{
\begin{tabular}{c||ccc|ccc}
\toprule
 & \multicolumn{3}{c|}{\textbf{FG Inpainting}} & \multicolumn{3}{c}{\textbf{BG Inpainting}} \\
\textbf{\sys} & FID $\uparrow$ & Prec $\downarrow$ & LPIPS $\uparrow$ & FID $\uparrow$ & Prec $\downarrow$ & LPIPS $\uparrow$ \\
\midrule
$m^{in}$ & 294.91 & 0.0044 & 0.6743 & 225.3 & 0.0024 & 0.5754 \\
$m^{out}$  & 292.98 & 0.0058 & 0.6813 & 258.43 & 0.0036 & 0.6249 \\
\bottomrule
\end{tabular}
}
\caption{Performance comparison of masks randomly shifted within the optimization boundary ($m^{in}$) and those exceeding the boundary ($m^{out}$).}
\label{table:masks}
\end{wraptable}

Figure~\ref{fig:masks} shows the inpainting results of our approach using diverse, boundary-exceeding masks $m^{out}$, where our method successfully protects the image from both foreground and background inpainting.
Table~\ref{table:masks} compares the quantitative results of our approach in both $m^{in}$ and $m^{out}$ mask settings. This demonstrates \sys's robust protection in real-world scenarios, maintaining strong performance even when user-defined masks exceed the optimization boundaries.

\vspace{-4mm}
\subsection{Discussion}
\vspace{-2mm}
\noindent\textbf{Transferability.} 
%
%
To demonstrate the transferability of our adversarial examples, we compare the effectiveness of \sys on image-to-image and text-to-image tasks using diffusion models with prior works in Appendix~\ref{appendix:transferability}. We observe that \sys exhibits comparable protections against image-to-image and text-to-image tasks, on par with the performance of prior methods that are solely designed for these tasks. We note that \sys is specifically designed for inpainting protection.

\noindent\textbf{Multi-object images.} We further evaluate the efficacy of \sys on images containing multiple objects by targeting the attention blocks for each object, as shown in Appendix~\ref{appendix:qual}. We first optimize perturbations within each object's mask $m$ and then apply perturbations to the remaining background. \sys remains effective regardless of the number of objects; however, the computational cost increases as the number of target objects increases. We leave addressing these computational overheads for future work.

\vspace{-3mm}
\section{Conclusion}
\vspace{-3mm}

In this paper, we present a novel image protection perturbation designed to defend against inpainting LDMs, which can replace masked regions with highly realistic objects or backgrounds. We are the first to bring attention to the dangers of inpainting tasks in image abuse and demonstrate the limitations of previous adversarial approaches in providing sufficient protection. To address the challenge of preventing malicious alterations in \textit{masked} regions with limited perturbations, \sys introduces attention disruption and a two-stage optimization strategy. By directly targeting the cross- and self-attention blocks, and optimizing separate perturbations for different object regions, \sys outperforms state-of-the-art methods in preventing inpainting manipulations. Additionally, \sys exhibits robustness to various hand-crafted masks, demonstrating its practical applicability in real-world scenarios.

\section*{Acknowledgements}
We would like to thank the anonymous reviewers for their constructive comments and suggestions. This work was supported by the National Research Foundation of Korea(NRF) grant funded by the Korea government(MSIT) (No. RS-2023-00208506) and the Institute of Information \& Communications Technology Planning \& Evaluation(IITP) grant funded by the Korea government (MSIT) (No. RS-2020-II200153, Penetration Security Testing of ML Model Vulnerabilities and Defense). Prof. Sung-Eui Yoon and Prof. Sooel Son are co-corresponding authors.

\bibliography{iclr2025_conference}
\bibliographystyle{iclr2025_conference}

\appendix
\newpage
\section{Appendix}

\subsection{Implementation Details}
\label{appendix:implementaion details}

\subsubsection{Algorithm of \sys}
\begin{algorithm}
\caption{\sys}
\begin{algorithmic}[1]
\State \textbf{Input:} Clean image $x$, perturbation $\delta$, mask set $M$, extracted feature $\phi$, optimization steps $N$, step size $\alpha$, perturbation budget $\eta$, total of $L$ layers in U-Net, timestep $T$
\State \textbf{Output:} Adversarial example $x'$
\State Initialize $\delta \sim \mathcal{U}(-\eta, \eta)$
\State $x' \gets x + \delta$
\For{mask \textbf{in} $M$}
    \State $m \gets \text{mask}$
    \State $x_0 \gets x \otimes m$
    \State $x'_0 \gets x' \otimes m$
    \For{$i = 0$ \textbf{to} $N-1$ \textbf{at timestep} $T$} \Comment{Optimization is performed at timestep $T$ only}
        \For{$l = 1$ \textbf{to} $L$}
            \vspace{0.5em}
            \State ($q_s^l, k_s^l, v_s^l) \gets (Q_s^l(\phi(x)), K_s^l(\phi(x)), V_s^l(\phi(x)))$
            \vspace{0.5em}
            \State $(q'{_s^l}, k'{_s^l}, v'{_s^l}) \gets (Q_s^{l}(\phi(x'_i)), K_s^l(\phi(x'_i)),V_s^l(\phi(x'_i)))$
            \vspace{0.5em}
            \State $ q_c^{l}, q'{_c^l} \gets Q_c^{l}(\phi(x)), Q_c^l(\phi(x'_i))$
            \vspace{0.5em}
        \EndFor
        \State $\mathcal{L}_{attn} \gets \sum\limits_{l} \big( \left\| q'{_s^l} - q_s^l \right\|^2 + \left\| k'{_s^l} - k_s^l \right\|^2  + \left\| v'{_s^l} - v_s^l \right\|^2 \big) + \sum\limits_{l} \big( \left\| q'{_c^l} - q_c^l \right\|^2$ \big)
        \State $\delta \gets \delta + \alpha \cdot \text{sign}(\nabla_{x'_i} \mathcal{L}_{attn})$
        \State $\delta \gets \text{clip}(\delta, -\eta, \eta)$
        \State $x'_{i+1} \gets x_0 + \delta$
    \EndFor
    \State $x' \gets x'_{N-1}$
\EndFor

\end{algorithmic}
\label{algo:advpaint}
\end{algorithm}

Algorithm~\ref{algo:advpaint} describes the perturbation generation process of \sys. Note that we optimize our perturbation only at timestep $T$, as considering additional timesteps significantly increase computational costs.

\subsubsection{Prior Adversarial Methods}
For all prior works used as our baselines~\citep{Photoguard, AdvDM, Mist, CAAT, SDST}, we follow their official implementations to optimize their perturbations. The only adjustment we made is to set the noise level by adjusting the hyperparameter $\eta$ to 0.06, ensuring that all methods operate under the same noise constraints. We note that all these baselines use PGD for optimizing their perturbations.

Several methods require setting a target latent for optimizing perturbations. For Photoguard~\citep{Photoguard}, we use the zero vector as the target latent, which is their default setting. For Mist~\citep{Mist} and SDST~\citep{SDST}, we use the target image of Mist for both implementations.

\subsubsection{Threat Model}
In this work, we evaluate our adversarial perturbations across a range of tasks, including inpainting, image-to-image, and text-to-image generation.

\noindent\textbf{Inpainting task:} We use the Stable Diffusion inpainting pipeline\footnote{https://huggingface.co/docs/diffusers/api/pipelines/stable\_diffusion/inpaint} provided by Diffusers (runwayml/stable-diffusion-inpainting). The default settings of the model are applied (inference step $T$=50, guidance scale=7.5, strength=1.0, etc.).

\noindent\textbf{Image-to-image task:} For image-to-image translation, we use the Stable Diffusion image-to-image pipeline\footnote{https://huggingface.co/docs/diffusers/api/pipelines/stable\_diffusion/img2img} provided by Diffusers (runwayml/stable-diffusion-v1-5). Specifically, we follow the default settings of the pipeline, where inference steps = 50, strength = 0.8, and guidance scale = 7.5.

\noindent\textbf{Text-to-image task:} We implement text-to-image generation using the \text{Textual Inversion}~\citep{textual-inversion} model, following the official implementation and settings from the paper. Specifically, we set the inference steps to 50 and the guidance scale to 7.5. For the input images, where 3 to 5 images are required, we utilized the official dataset of DreamBooth~\citep{DreamBooth}. For both tasks, we used images of 512×512 size and randomly crafted the conditional prompts.

\subsubsection{Generating Prompts for Optimization and Inpainting}
In the process of generating adversarial perturbations using \sys, our target inpainting model requires a prompt input as an external condition. For simplicity, we manually set the prompt as a basic $\{\text{noun}\}$ format (\eg ``A gorilla" for gorilla images, ``A dog" for dog images).

In the inference phase, as described in~\ref{Experimental Setup}, we generated 50 random prompts using ChatGPT~\citep{ChatGPT}. 
For foreground inpainting, the prompts followed the format of $\{\text{noun}\}$ (\eg ``An orange", ``A tiger"). For background inpainting,  we generated prompts in the format of $\{\text{preposition}, \text{location}\}$ (\eg ``at the riverside.", ``at a wooden fence."), inserting the prompts used in the perturbation-generation step at the beginning of each generated prompt (\eg ``A gorilla at the riverside.", ``A dog at a wooden fence."). We followed the prompt setup from \cite{inpaintanything} for fair comparisons.

\subsection{U-Net Denoiser Modified for Inpainting}
\label{appendix:U-Net}

\begin{figure}  
    \begin{center}
    \includegraphics[width=0.95\linewidth, trim={0.5cm 4.1cm 0.5cm 5.7cm}, clip]{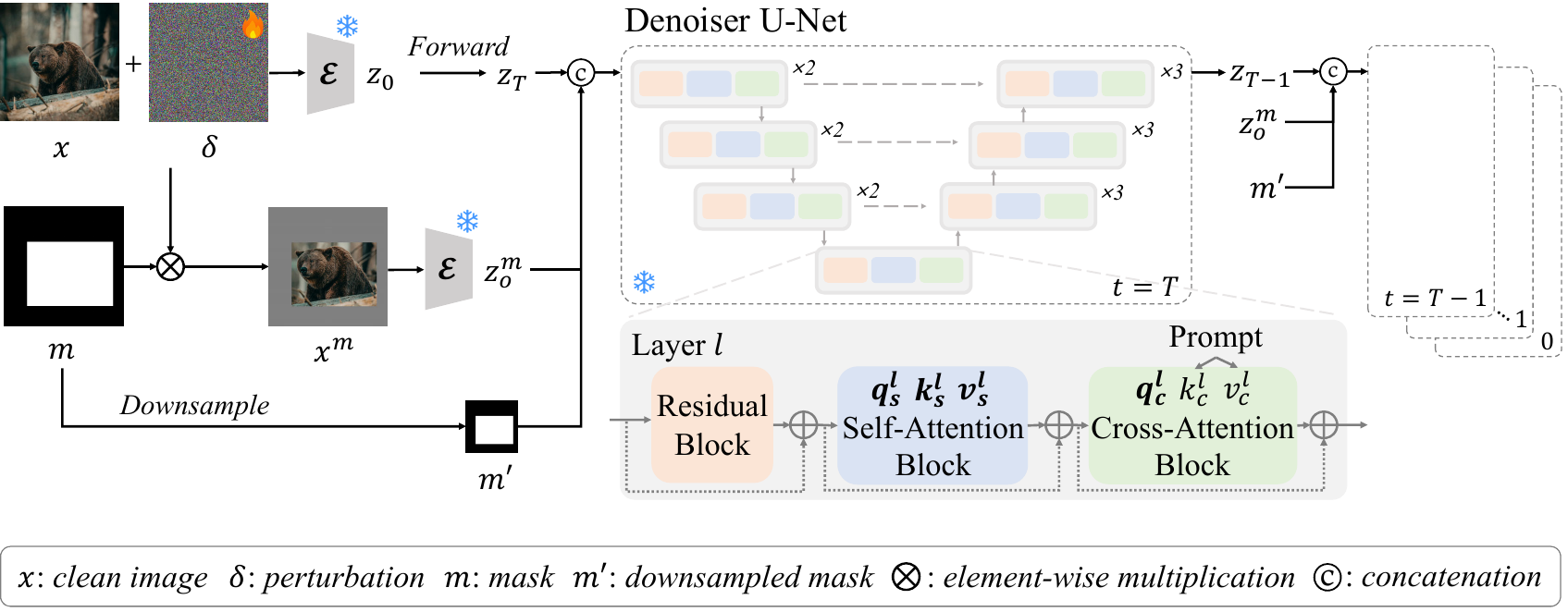}
    \end{center}
    
    \caption{The architecture of the LDM denoiser specifically modified for inpainting tasks. The input image $x$ and the masked image $x^m$ share the same encoder $\varepsilon$. The latent $z_0^{m}$ and the resized mask $m'$ are fed into the model at every timestep. Here, $c$ denotes from cross-attention and $s$ stands for self-attention. We optimize the perturbation $\delta$ by targeting the bolded components in each block of each layer $l$.}
   \label{fig:architecture}
\end{figure}

\subsubsection{Preliminary: Attention Blocks in LDMs}
\label{sec:att_block}
\cite{LDM} proposed an LDM that leverages self- and cross-attention blocks. The self-attention blocks play a crucial role in generating high-dimensional images by capturing long-range dependencies between spatial regions of input images. Meanwhile, the cross-attention blocks are designed to align the latent image representation with external inputs, such as prompts, during the denoising process, ensuring that the generated image reflects the desired conditioning~\citep{p2p, PnP, AttnLayersinSD}. 

\subsubsection{Architecture of inpainting LDM}
\label{appendix:architecture}
We demonstrate the architecture of inpainting LDM in Figure~\ref{fig:architecture}. This model takes three inputs: the original image $x$, the mask $m$, and the masked image $x^m = x \otimes m$, where $\otimes$ represents element-wise multiplication. 
The shared encoder $\mathcal{E}$ produces two latent vectors, $z_0$ and $z_0^m$. 

The denoiser U-Net consists of 16 layers, each comprising a sequence of residual, self-attention, and cross-attention blocks, along with skip connections. As in the default LDM, the denoiser predicts the noise added to the latent and denoises the latent $z_t$ at each timestep $t$. The resulting denoised latent $z_{t-1}$ is then concatenated with $z_0^m$ and $m'$ for the next denoising step. Note that 
After the denoising for inference steps $T$, the denoised latent $z'_0$ is then inserted to the same decoder $\mathcal{D}$ of default LDM to generate the inpainted image.


\subsection{Transferability of \sys}
\label{appendix:transferability}
\subsubsection{Variants of inpainting models}

\begin{table}
\huge
\centering
\resizebox{\textwidth}{!}{
\begin{tabular}{c||ccc|ccc|ccc|ccc}
\toprule
 & \multicolumn{6}{c|}{\textbf{Foreground Inpainting}} & \multicolumn{6}{c}{\textbf{Background Inpainting}} \\
\cmidrule(lr){2-7} \cmidrule(lr){8-13}
\textbf{HD-Painter} & \multicolumn{3}{c|}{$\boldsymbol{m^{seg}}$} & \multicolumn{3}{c|}{$\boldsymbol{m^{bb}}$} & \multicolumn{3}{c|}{$\boldsymbol{m^{seg}}$} & \multicolumn{3}{c}{$\boldsymbol{m^{bb}}$} \\
 & FID $\uparrow$ & Prec $\downarrow$   & LPIPS $\uparrow$  & FID $\uparrow$  & Prec $\downarrow$  & LPIPS $\uparrow$  & FID $\uparrow$  & Prec $\downarrow$   & LPIPS $\uparrow$  & FID $\uparrow$  & Prec $\downarrow$  & LPIPS $\uparrow$  \\
\midrule
Photoguard & 153.46 & 0.8552 & 0.5632 & 132.63 & 0.8962 & 0.5446 & 93.46 & 0.5978 & 0.3064 & 127.90 & 0.3246 & 0.4400 \\
AdvDM & 155.44 & 0.6180 & 0.4807 & 134.54 & 0.7032 & 0.4707 & 75.85 & 0.7278 & 0.2538 & 109.28 & 0.4738 & 0.3617 \\
SDST & 146.85 & 0.8568 & 0.4462 & 128.88 & 0.9038 & 0.4456 & 87.64 & 0.6042 & 0.2896 & 127.85 & 0.3366 & 0.4120 \\
 \midrule
\textbf{\sys} & \textbf{178.71} & \textbf{0.5350} & \textbf{0.5770} & \textbf{156.51} & \textbf{0.6276} & \textbf{0.5754} & \textbf{164.10} & \textbf{0.3310} & \textbf{0.3998} & \textbf{264.79} & \textbf{0.1748}& \textbf{0.5232} \\

\bottomrule
\end{tabular}
}
\caption{Quantitative comparison for HD-Painter~\citep{HD-Painter}.}
\label{table:HD-Painter}
\end{table}

\begin{table}
\huge
\centering
\resizebox{\textwidth}{!}{
\begin{tabular}{c||ccc|ccc|ccc|ccc}
\toprule
 & \multicolumn{6}{c|}{\textbf{Foreground Inpainting}} & \multicolumn{6}{c}{\textbf{Background Inpainting}} \\
\cmidrule(lr){2-7} \cmidrule(lr){8-13}
\textbf{DreamShaper} & \multicolumn{3}{c|}{$\boldsymbol{m^{seg}}$} & \multicolumn{3}{c|}{$\boldsymbol{m^{bb}}$} & \multicolumn{3}{c|}{$\boldsymbol{m^{seg}}$} & \multicolumn{3}{c}{$\boldsymbol{m^{bb}}$} \\
 & FID $\uparrow$ & Prec $\downarrow$   & LPIPS $\uparrow$  & FID $\uparrow$  & Prec $\downarrow$  & LPIPS $\uparrow$  & FID $\uparrow$  & Prec $\downarrow$   & LPIPS $\uparrow$  & FID $\uparrow$  & Prec $\downarrow$  & LPIPS $\uparrow$  \\
\midrule
Photugard & 188.08 & 0.7422 & 0.5878 & 157.90 & 0.8544 & 0.5792 & 103.74 & 0.6112 & 0.3361 & 131.61 & 0.3074 & 0.4840 \\
AdvDM & 183.74 & 0.5114 & 0.5080 & 152.55 & 0.6540 & 0.5001 & 84.99 & 0.7102 & 0.2846 & 115.41 & 0.4340 & 0.3992 \\
SDST & 179.80 & 0.7792 & 0.4682 & 151.38 & 0.8656 & 0.4725 & 98.67 & 0.5954 & 0.3149 & 132.79 & 0.2978 & 0.4446 \\
 \midrule
\textbf{\sys} & \textbf{230.53} & \textbf{0.3856} & \textbf{0.6042} & \textbf{186.27} & \textbf{0.5196} & \textbf{0.6092} & \textbf{177.68} & \textbf{0.3160} & \textbf{0.4317} & \textbf{266.85} & \textbf{0.1622} & \textbf{0.5561} \\

\bottomrule
\end{tabular}
}
\caption{Quantitative comparison for DreamShaper~\citep{dreamshaper2024}.}
\label{table:DreamShaper}
\end{table}

\begin{table}[t!]
\huge
\centering
\resizebox{\textwidth}{!}{
\begin{tabular}{c||ccc|ccc|ccc|ccc}
\toprule
 & \multicolumn{6}{c|}{\textbf{Foreground Inpainting}} & \multicolumn{6}{c}{\textbf{Background Inpainting}} \\
\cmidrule(lr){2-7} \cmidrule(lr){8-13}
\textbf{SD-2-Inp.} & \multicolumn{3}{c|}{$\boldsymbol{m^{seg}}$} & \multicolumn{3}{c|}{$\boldsymbol{m^{bb}}$} & \multicolumn{3}{c|}{$\boldsymbol{m^{seg}}$} & \multicolumn{3}{c}{$\boldsymbol{m^{bb}}$} \\
 & FID $\uparrow$ & Prec $\downarrow$   & LPIPS $\uparrow$  & FID $\uparrow$  & Prec $\downarrow$  & LPIPS $\uparrow$  & FID $\uparrow$  & Prec $\downarrow$   & LPIPS $\uparrow$  & FID $\uparrow$  & Prec $\downarrow$  & LPIPS $\uparrow$  \\
\midrule
Photoguard & 239.73 & 0.5102 & 0.6226 & 199.21 & 0.7000 & 0.6071 & 110.33 & 0.4570 & 0.3798 & 126.74 & 0.1712 & 0.5094 \\
AdvDM & 249.57 & 0.1902 & 0.5393 & 199.22 & 0.3636 & 0.5246 & 89.66 & 0.5942 & 0.3027 & 114.39 & 0.2610 & 0.4197 \\
SDST & 231.96 & 0.5324 & 0.5001 & 201.65 & 0.6756 & 0.4996 & 106.61 & 0.4718 & 0.3569 & 130.00 & 0.1892 & 0.4710 \\

 \midrule
\textbf{\sys} &\textbf{325.14} & \textbf{0.0926} & \textbf{0.6452} & \textbf{264.72} & \textbf{0.2160} & \textbf{0.6443} & \textbf{198.32} & \textbf{0.2210} & \textbf{0.4633} & \textbf{267.91} & \textbf{0.0842} & \textbf{0.5756} \\

\bottomrule
\end{tabular}
}
\caption{Quantitative comparison for Stable-Diffusion-2-Inpainting model.}
\label{table:SD2}
\end{table}

We conducted extensive experiments on multiple inpainting model variants: HD-Painter~\citep{HD-Painter}, DreamShaper~\citep{dreamshaper2024}, and the Stable Diffusion v2 inpainting model~\citep{stablediffusion2inpainting}. For all the variants, we follow the default settings of the official code and we also follow the default settings in the paper, only replacing the inpainting model to one of the variants.

\begin{figure}[b]
    \begin{center}
    \includegraphics[width=0.8\linewidth, trim={2cm 8cm 2cm 8cm}, clip]{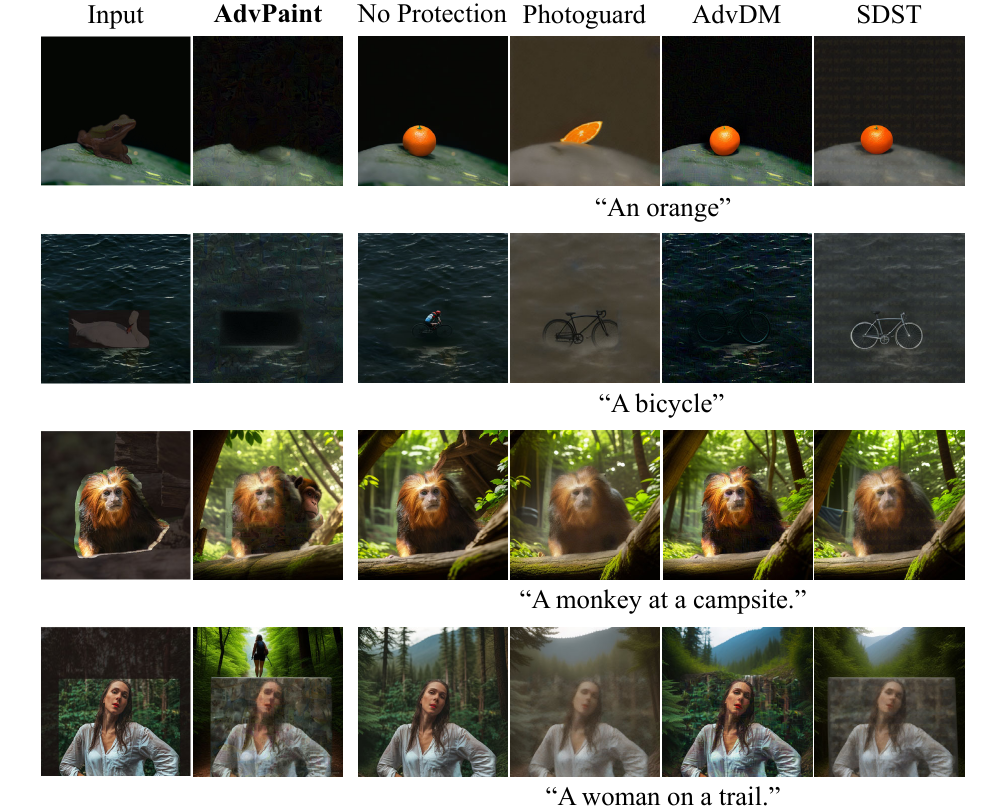}
    \end{center}
    \caption{Qualitative inpainting results of HD-Painter~\citep{HD-Painter}.}
   \label{fig:HD-Painter}
\end{figure}

\begin{figure}  
    \begin{center}
    \includegraphics[width=0.8\linewidth, trim={2cm 8cm 2cm 8cm}, clip]{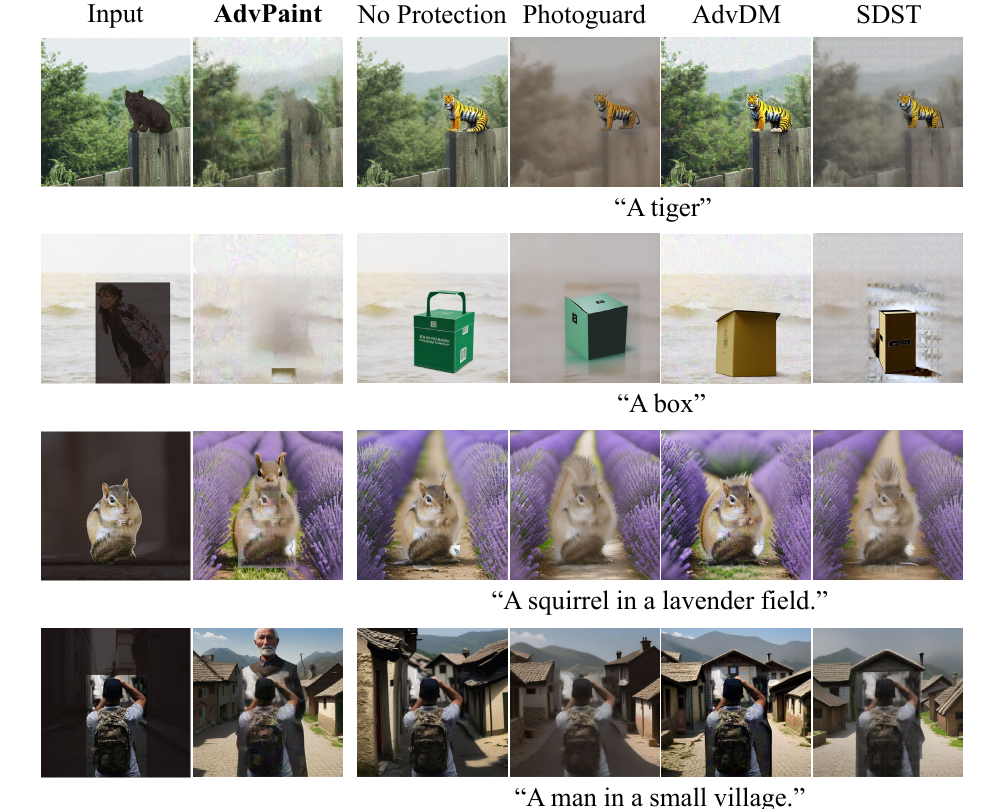}
    \end{center}
    \caption{Qualitative inpainting results of DreamShaper~\citep{dreamshaper2024}.}
   \label{fig:DreamShaper}
\end{figure}

\begin{figure}  
    \begin{center}
    \includegraphics[width=0.8\linewidth, trim={2cm 8cm 2cm 8cm}, clip]{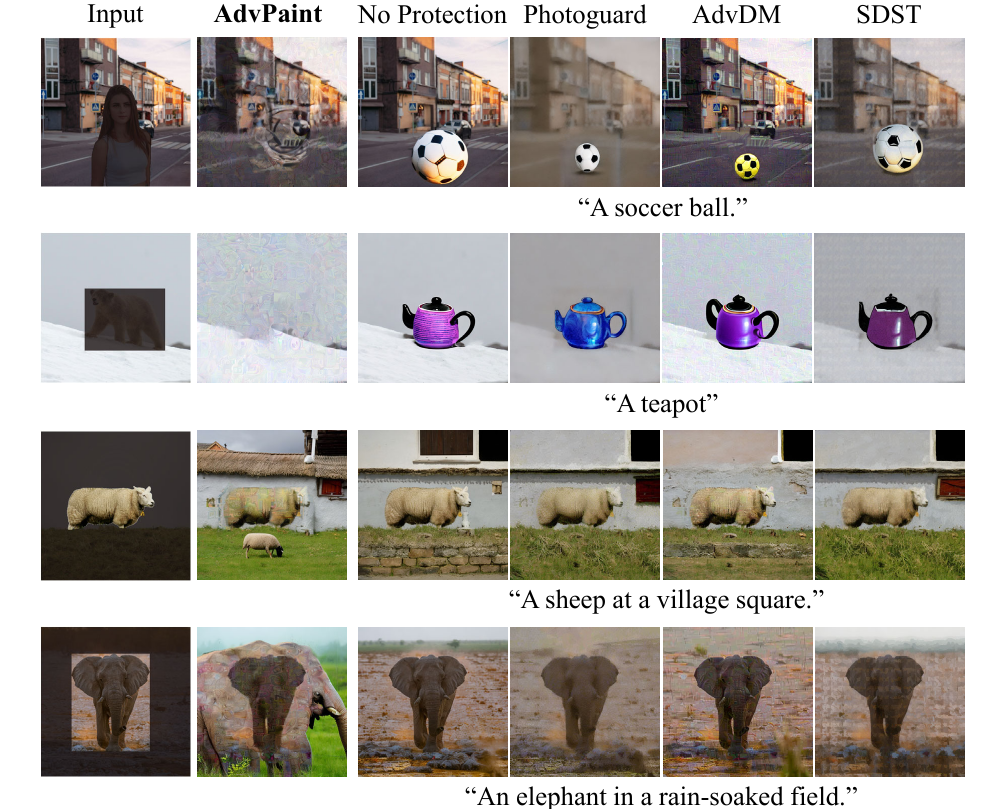}
    \end{center}
    \caption{Qualitative inpainting results of Stable-Diffusion-2-Inpainting model.}
   \label{fig:SD2}
\end{figure}

In Table~\ref{table:HD-Painter},~\ref{table:DreamShaper}, and~\ref{table:SD2}, we evaluate \sys against these variants using FID, precision, and LPIPS metrics and compared it with other baseline protection methods. Even when the architecture differed significantly (e.g., HD-Painter) or when fine-tuning changed the model parameters (e.g., DreamShaper, SD-2-inpainting), \sys consistently outperform earlier protection methods across all metrics. Qualitative results are depicted in Figure~\ref{fig:HD-Painter},~\ref{fig:DreamShaper}, and~\ref{fig:SD2}.

\subsubsection{Image-to-image and text-to-image tasks}
\label{sec:i2i}
\begin{figure}[t!]
    \begin{center}
    \includegraphics[width=0.95\linewidth, trim={1.2cm 1.4cm 1.2cm 1.4cm}, clip]{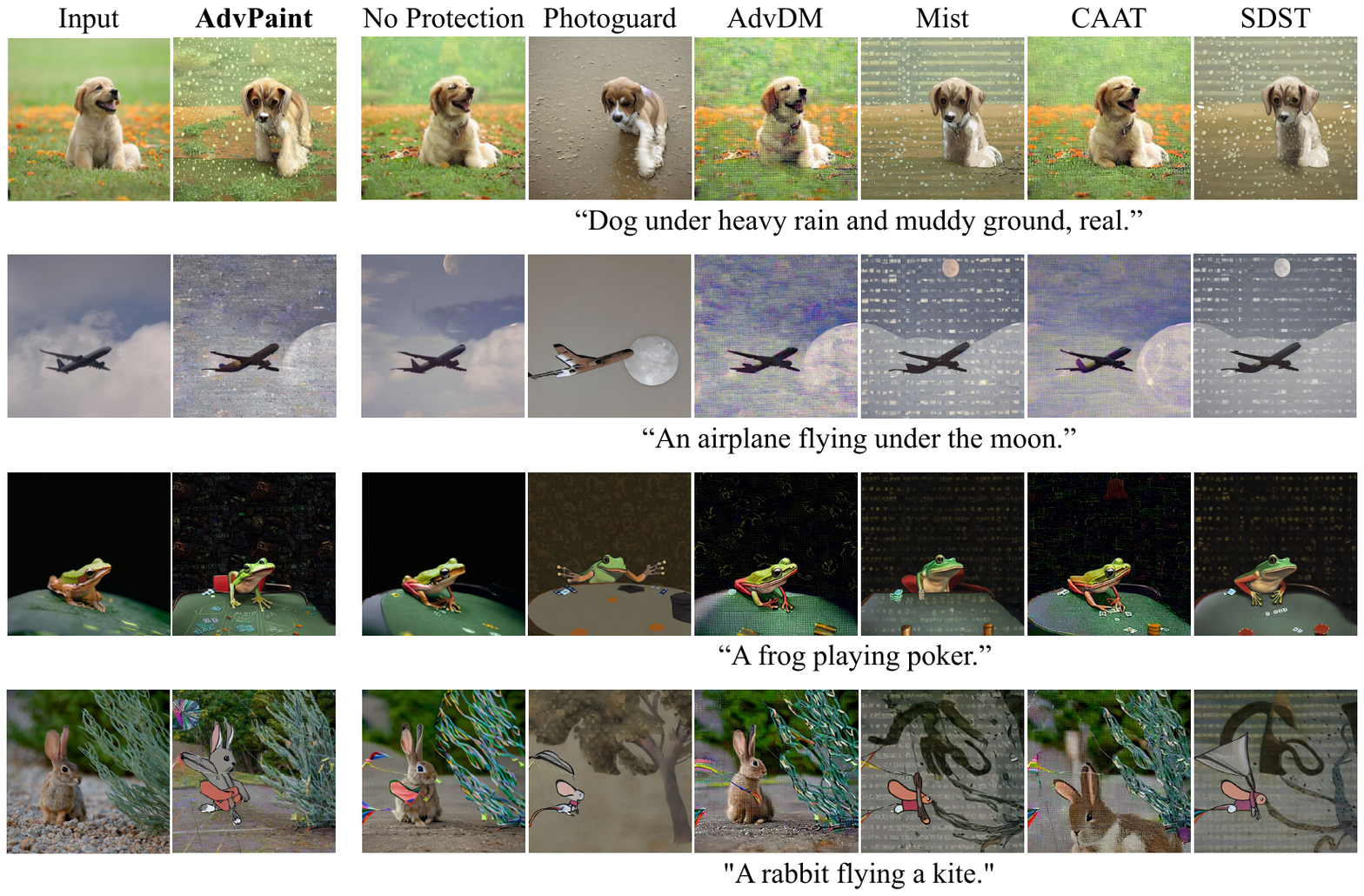}
    \end{center}
    \caption{Comparison in image-to-image translation task. The results are generated via Stable Diffusion image-to-image pipeline.}
   \label{fig:i2i}
\end{figure}

\begin{figure}  
    \begin{center}
    \includegraphics[width=0.95\linewidth, trim={1.2cm 3.5cm 1.2cm 3.5cm}, clip]{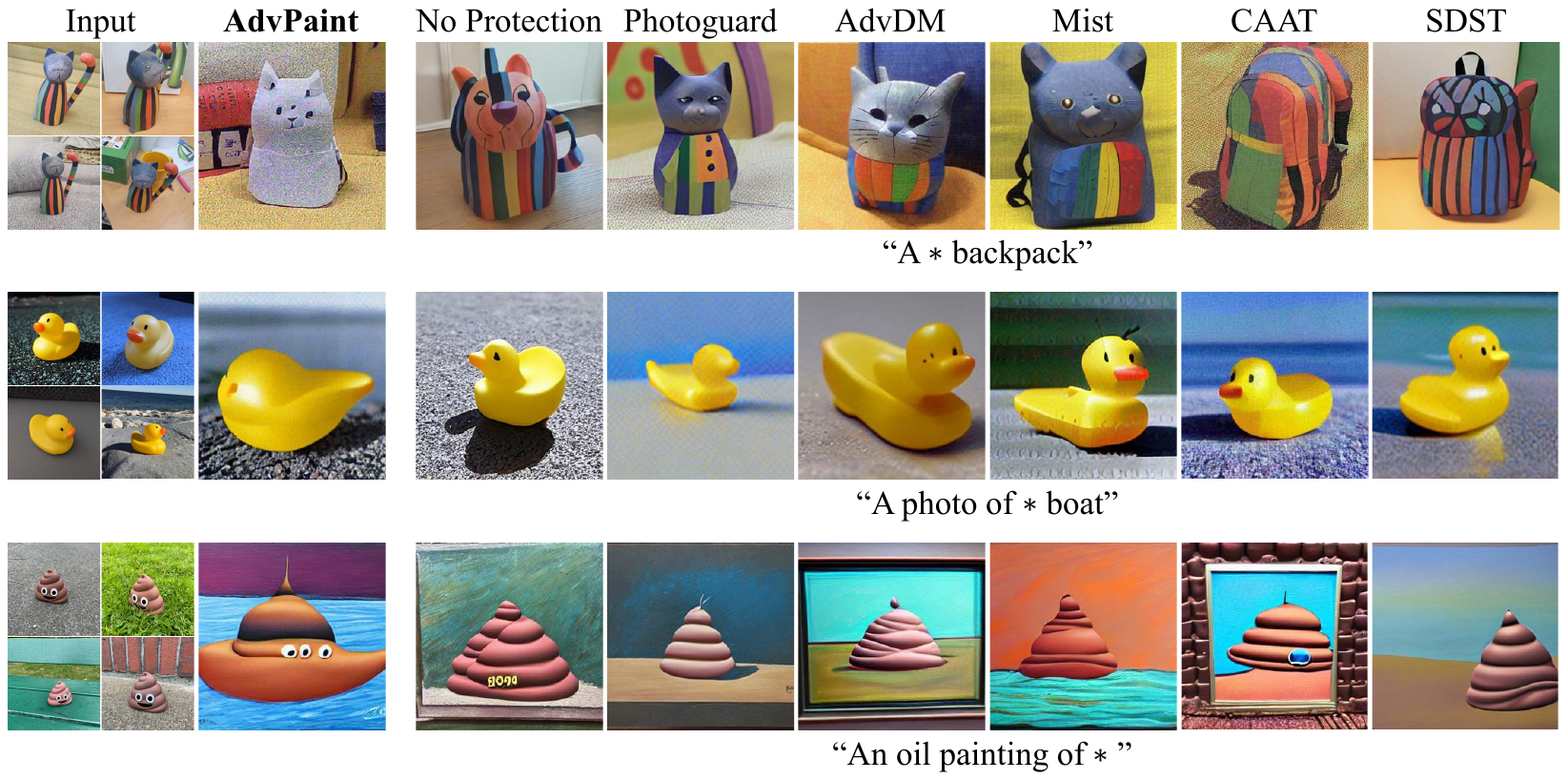}
    \end{center}
    \caption{Comparison of text-to-image generation. The $*$ in the prompts indicates the representative prompt corresponding to the input images. The results are generated via Textual Inversion~\citep{textual-inversion}.}
   \label{fig:t2i}
\end{figure}

We demonstrate the transferability of \sys to image-to-image and text-to-image tasks in Figure~\ref{fig:i2i} and~\ref{fig:t2i}. While prior methods targeting these tasks effectively protect images from manipulations, our approach also delivers competitive safeguarding results.

\subsubsection{DiT-based generation models}
DiT~\citep{DiT} suggests a new paradigm in text-to-image generation tasks by applying vision transformers to Latent Diffusion Models, which decreases model complexity and increases generation quality. we evaluated the robustness of \sys against an adversary using the inpainting model of Flux~\citep{Flux} and Stable Diffusion 3 (SD3)~\citep{SD3}, and text-to-image model Pixart-$\delta$~\citep{Pixart-delta} which leverages a diffusion transformer. Unlike Pixart-$\delta$, we note that models like DiT and Pixart-$\alpha$~\citep{Pixart-alpha} are designed for generating images solely from text prompts using diffusion transformer architectures, which make them unsuitable for our tasks that require accepting input images.

\textbf{Flux} provides an inpainting module based on multi-modal and parallel diffusion transformer blocks. We utilized the “black-forest-labs/FLUX.1-schnell” checkpoint and the image size was set to 512x512 to match our settings.

\begin{figure}  
    \begin{center}
    \includegraphics[width=0.95\linewidth, trim={2cm 4cm 2cm 4cm}, clip]{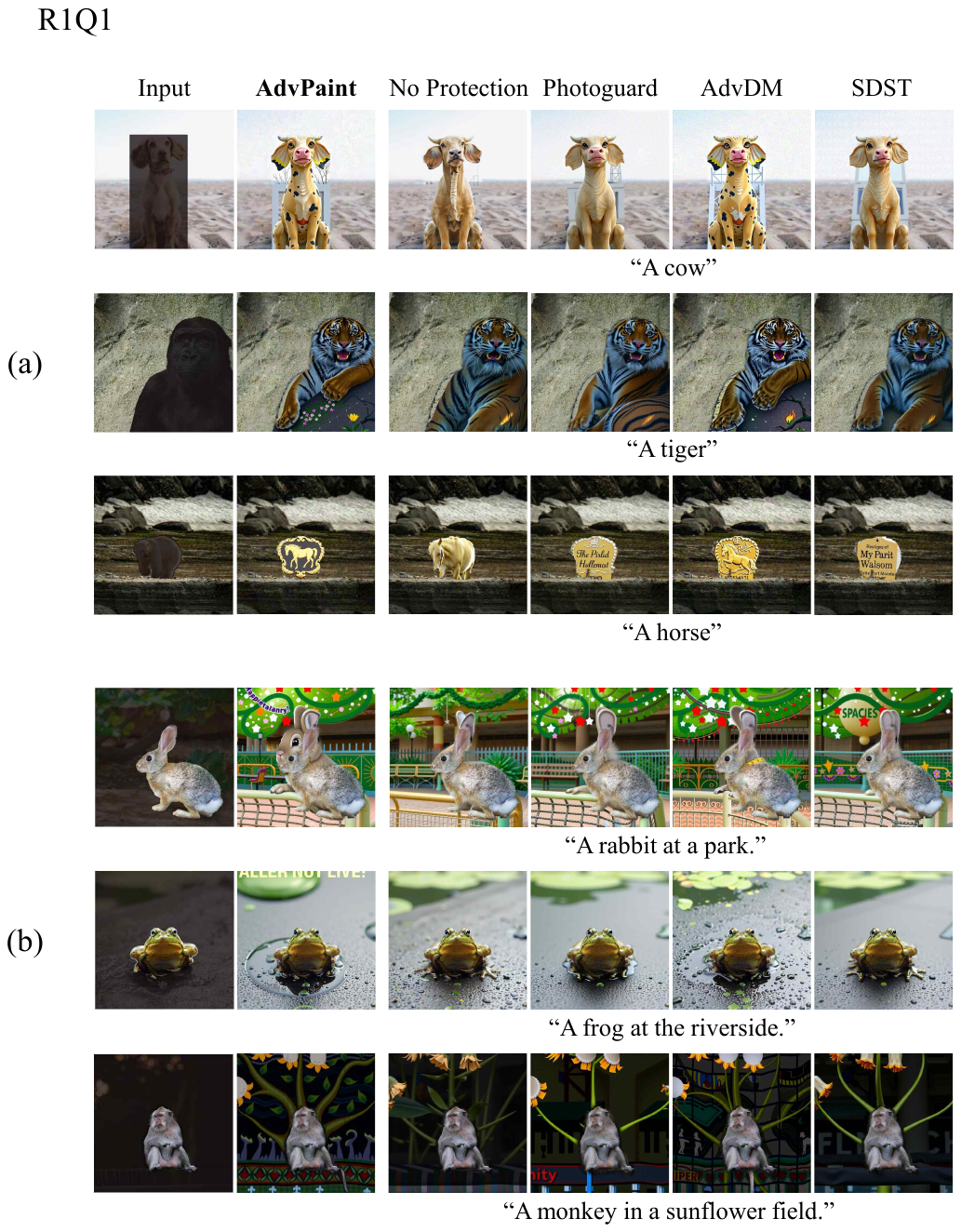}
    \end{center}
    \caption{Qualitative results of \sys and baseline models applied to Flux~\citep{Flux}. Results demonstrate the transferability of \sys to DiT-based inpainting models, causing misalignment between generated regions and unmasked areas in both (a) foreground and (b) background inpainting tasks. Dark parts in the input image indicate the masked regions.}
   \label{fig:Flux}
\end{figure}

As shown in Figure~\ref{fig:Flux}, \sys effectively disrupts the inpainting process by causing misalignment between generated regions and unmasked areas. For example, it generates cartoon-style cows in (a) and adds a new rabbit in (b), while also producing noisy patterns in the unmasked areas of the images.

\textbf{SDS} is a text-to-image model built on the architecture of a Multi-modal DiT (MMDiT) and includes an inpainting pipeline, making it suitable for our experiments. We followed the official implementation, modifying only the image size to 512x512 to match our experimental settings.

\begin{figure}  
    \begin{center}
    \includegraphics[width=0.95\linewidth, trim={2cm 4cm 2cm 4cm}, clip]{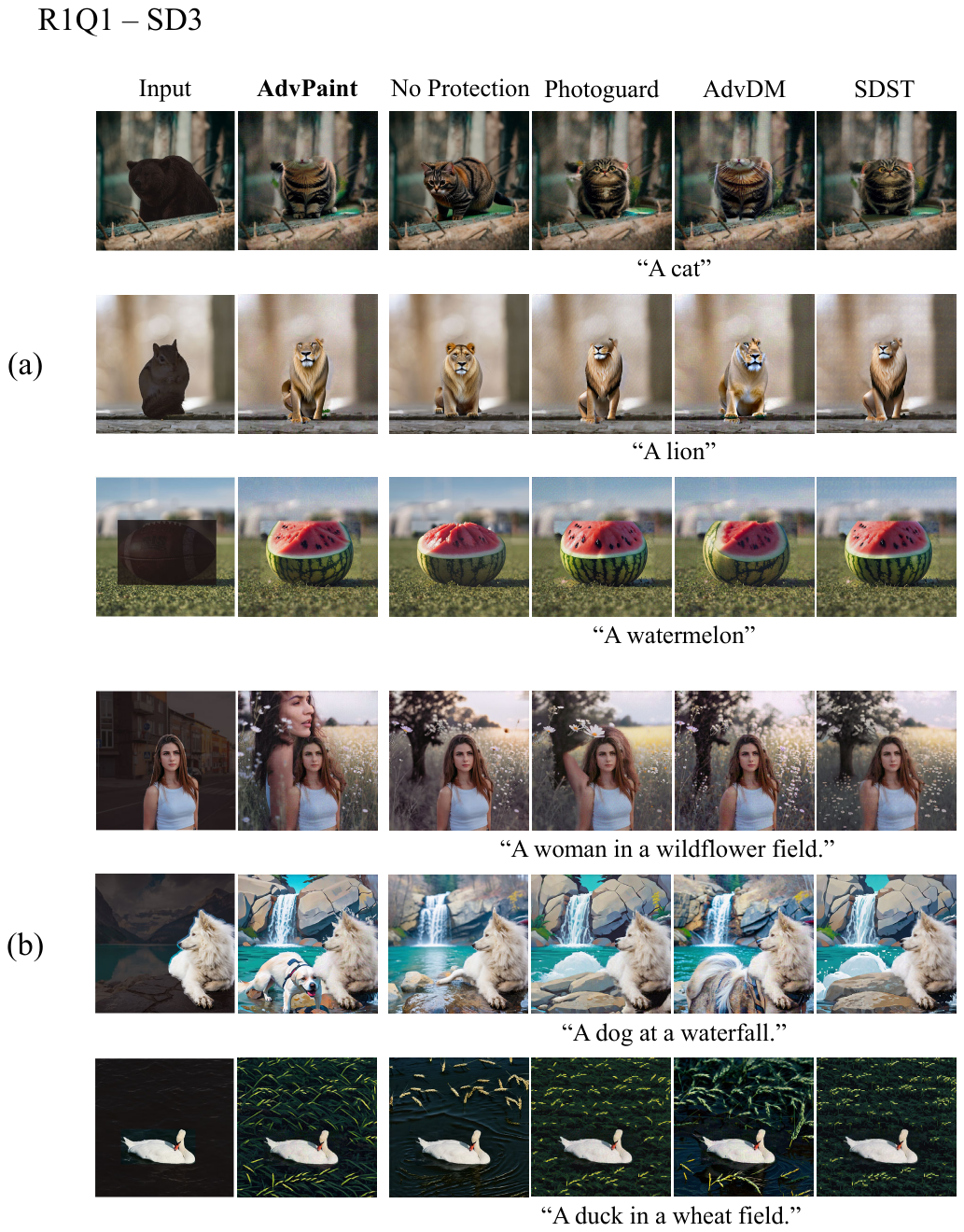}
    \end{center}
    \caption{Qualitative results of \sys and baseline models applied to SD3~\citep{SD3}. Results demonstrate the transferability of \sys to DiT-based inpainting models, causing misalignment between generated regions and unmasked areas in both (a) foreground and (b) background inpainting tasks. Dark parts in the input image indicate the masked regions.}
   \label{fig:SD3}
\end{figure}

As shown in the updated Figure~\ref{fig:SD3}, our results demonstrate the protective capabilities of AdvPaint against DiT-based inpainting tasks. Notably, we observed misalignment between generated images and unmasked regions. For instance, parts of a cat, lion, and watermelon are not fully generated and appear hidden behind the unmasked region in (a). In (b), which involves background inpainting tasks, the backgrounds are cartoonized, often disregarding pre-existing objects and generating new ones. This protective effect, which disrupts the semantic connection with unmasked objects, is also evident in the results for Flux.

\textbf{Chen et al.} have proposed Pixart-$\delta$ which incorporates DreamBooth~\citep{DreamBooth} into DiT. We chose this work for the adversary’s generative model since it supports feeding an input image along with a command prompt for performing generation. 

\begin{figure}  
    \begin{center}
    \includegraphics[width=0.9\linewidth, trim={1.5cm 7.2cm 1.5cm 7.2cm}, clip]{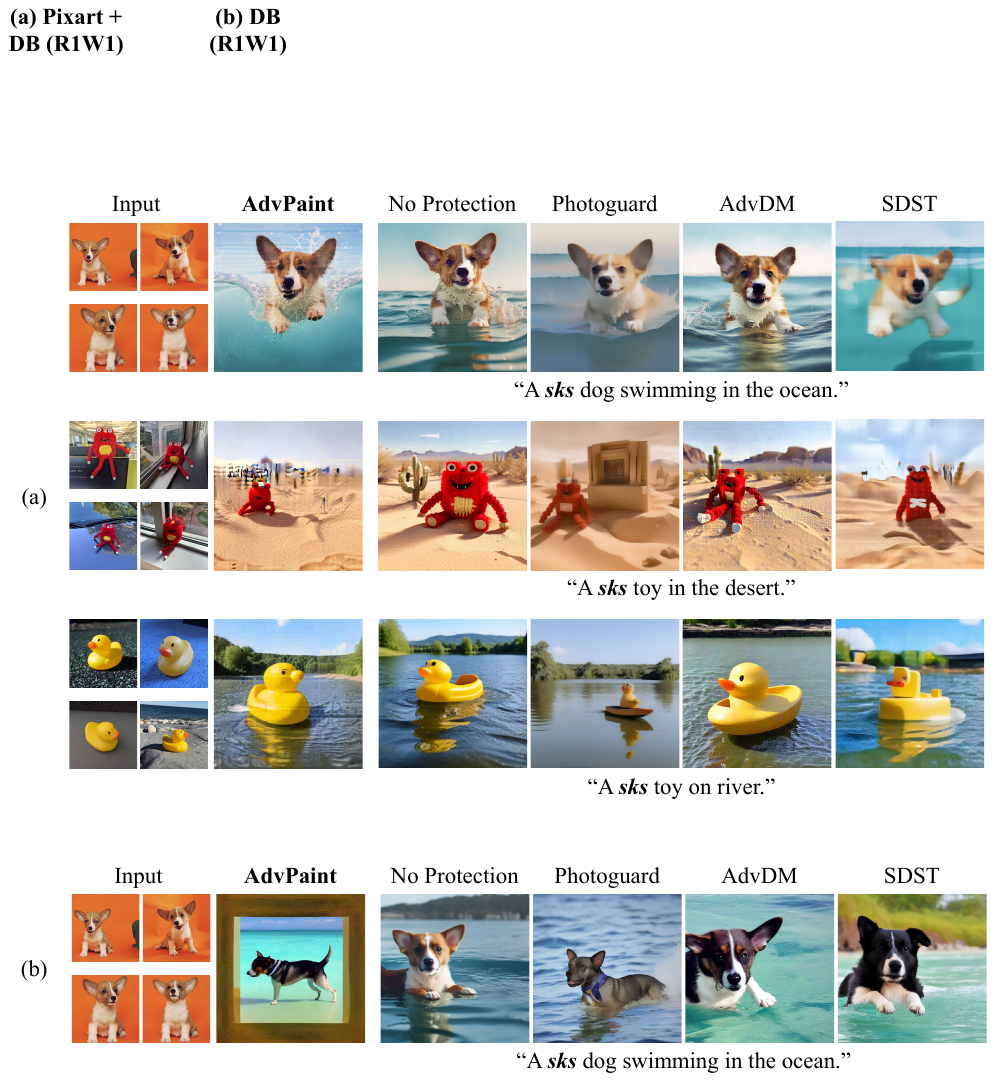}
    \end{center}
    \caption{Comparison of text-to-image generation from (a) DiT-based DreamBooth~\citep{Pixart-delta} and (b) the original DreamBooth~\citep{DreamBooth}.}
   \label{fig:DiT}
\end{figure}

As shown in Figure~\ref{fig:DiT} (a), AdvPaint-generated perturbations consistently undermine the generation ability of Pixart-$\delta$. Furthermore, AdvPaint also renders noise patterns that degrade the image quality on the resulting output images of the diffusion model, which aligns with the behavior of previous methods (\ie~Photoguard, AdvDM, SDST).

\sys also effectively disrupts the original DreamBooth~\citep{DreamBooth}, as shown in Figure~\ref{fig:DiT} (b). However, our findings indicate that \sys and the previous methods are less effective against Pixart-$\delta$ that leverages DiT, as shown in Figure~\ref{fig:DiT} (a). Additionally, compared to the results of LDM-based inpainting models in Figure~\ref{fig:prior works}, current methods are less effective when applied to DiTs. Discernible objects are generated in the foreground inpainting tasks and new objects according to the prompts are not always generated. We believe this ineffectiveness stems from the distinct characteristic of DiT, which processes patchified latent representations. \sys and our baselines are specifically designed to target LDMs, which utilize the entire latent representation as input, allowing perturbations to be optimized over the complete latent space. Thus, when latents are patchified in DiTs, perturbations may become less effective at disrupting the model's processing, thereby diminishing their protective capability. This discrepancy necessitates further research to develop protection methods specifically tailored to safeguard images against the adversary misusing DiT-based models. For instance, optimizing perturbations at the patch level rather than across the entire latent representation could prove more effective in countering the unique paradigm of image generation in DiT-based models.

\subsection{Different prompts for various inpainting Tasks}
In real-world scenarios, the exact prompts used by adversaries to maliciously modify images remain unknown. To simulate and analyze potential attack vectors, we conduct experiments using a diverse set of prompts that are likely candidates for foreground and background inpainting tasks. 

Please note that below experiments were conducted under the same default settings (\ie~using the Stable Diffusion Inpainting model with a total of 100 images and 50 prompts per image), ensuring a fair and consistent comparison.

\subsubsection{Foreground Inpainting}
In the experiments throughout the paper, prompts follow the format of $\{\text{noun}\}$ for foreground inpainting tasks. Here, we evaluate the robustness of \sys using a different kind of prompt: a prompt that describes the \textit{mask-covered} object itself. For example, we used the prompt “A man” for an input image describing a male and performed an inpainting task to generate another male image.

\begin{figure}  
    \begin{center}
    \includegraphics[width=0.77\linewidth, trim={2cm 7.9cm 2cm 7.9cm}, clip]{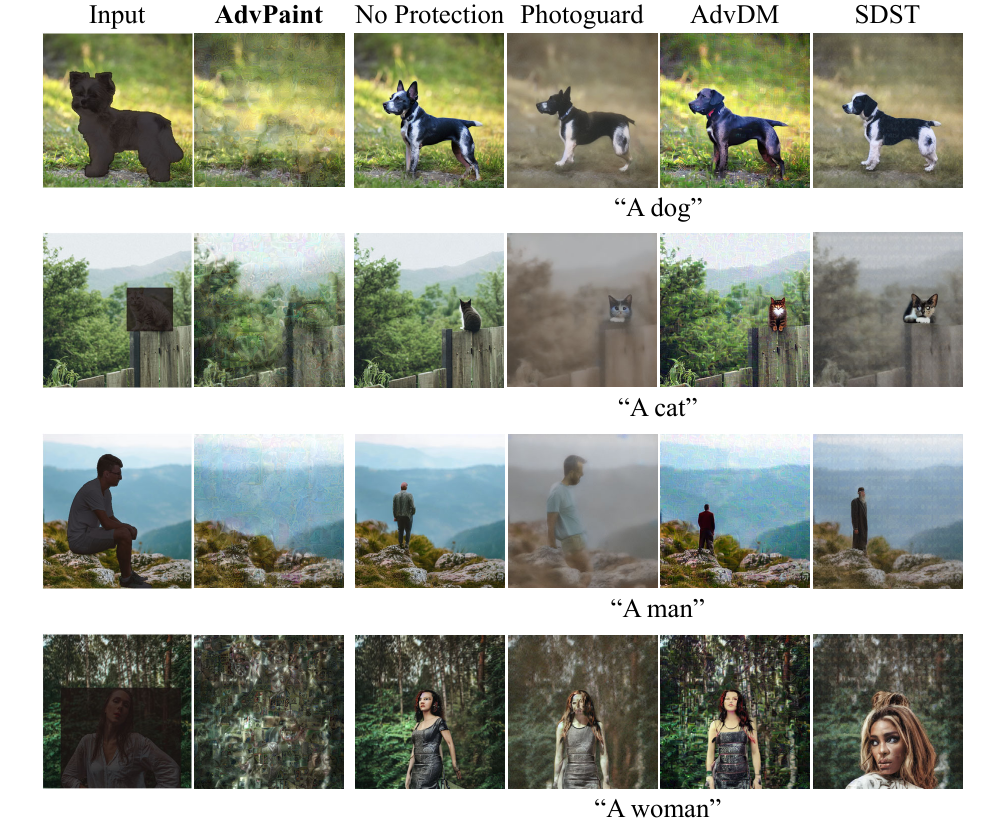}
    \end{center}
    \caption{Qualitative results of foreground inpainting with prompts that describe the mask-covered object. Dark parts in the input image indicate the masked regions.}
   \label{fig:R1Q3}
\end{figure}

As demonstrated in Table~\ref{table:R1Q3Q4} (a), \sys successfully disrupted the adversary's inpainting task, resulting in the generation of an image with no discernible object. This is because the perturbation optimized to disrupt the attention mechanism successfully redirects the attention to other unmasked areas as explained in Section~\ref{sec:exp_att} and Figure~\ref{fig:attmaps}. We demonstrate the qualitative examples in Figure~\ref{fig:R1Q3}.

\begin{table}
\huge
\centering
\resizebox{\textwidth}{!}{
\begin{tabular}{c||ccc|ccc|ccc|ccc}
\toprule
 & \multicolumn{6}{c|}{\textbf{(a) FG Inpainting}} & \multicolumn{6}{c}{\textbf{(b) BG Inpainting}} \\
\cmidrule(lr){2-7} \cmidrule(lr){8-13}
 & \multicolumn{3}{c|}{$\boldsymbol{m^{seg}}$} & \multicolumn{3}{c|}{$\boldsymbol{m^{bb}}$} & \multicolumn{3}{c|}{$\boldsymbol{m^{seg}}$} & \multicolumn{3}{c}{$\boldsymbol{m^{bb}}$} \\
 Optim. Methods & FID $\uparrow$ & Prec $\downarrow$   & LPIPS $\uparrow$  & FID $\uparrow$  & Prec $\downarrow$  & LPIPS $\uparrow$  & FID $\uparrow$  & Prec $\downarrow$   & LPIPS $\uparrow$  & FID $\uparrow$  & Prec $\downarrow$  & LPIPS $\uparrow$  \\
\midrule
 Photoguard & 161.44 & 0.0874 & 0.6415 & 129.99 & 0.2158 & 0.6171 & 144.21 & 0.5230 & 0.4063 & 153.63 & 0.2280 & 0.5317 \\
 AdvDM & 160.54 & 0.0658 & 0.5167 & 127.36 & 0.1266 & 0.5122 & 118.13 & 0.6228 & 0.3168 & 131.58 & 0.2720 & 0.4311 \\
 SDST & 148.57 & 0.1340 & 0.4930 & 120.55 & 0.2456 & 0.4882 & 139.86 & 0.5112 & 0.3810 & 152.73 & 0.2280 & 0.4892 \\
 \midrule
 \sys & \textbf{331.27} & \textbf{0.0036} & \textbf{0.6706} & \textbf{275.48} & \textbf{0.0264} & \textbf{0.6697} & \textbf{291.12} & \textbf{0.3490} & \textbf{0.4948} & \textbf{355.94} & \textbf{0.1152} & \textbf{0.6014} \\
\bottomrule
\end{tabular}
}
\caption{Quantitative comparison with a diverse set of prompts that are likely candidates for foreground and background inpainting tasks. We set prompts as (a) $\{\text{noun}\}$ that describes the \textit{mask-covered} object and (b) $\{\text{preposition}, \text{location}\}$.}
\label{table:R1Q3Q4}
\end{table}

\subsubsection{Background Inpainting}
\begin{figure}  
    \begin{center}
    \includegraphics[width=0.95\linewidth, trim={2cm 4cm 2cm 4cm}, clip]{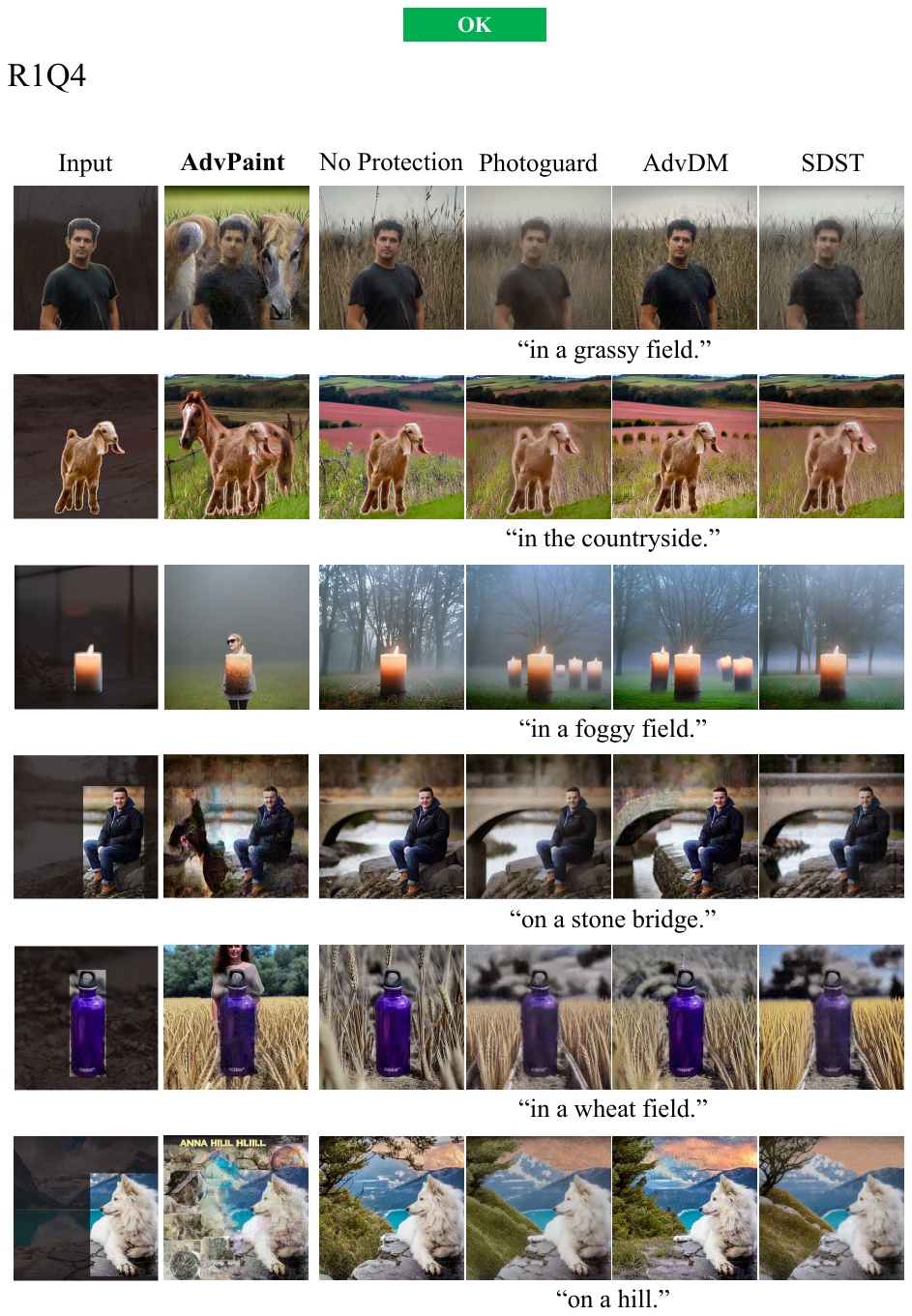}
    \end{center}
    \caption{Qualitative results of background inpainting with prompts that follow the format of $\{\text{preposition}, \text{location}\}$. Dark parts in the input image indicate the masked regions.}
   \label{fig:R1Q4}
\end{figure}

For background tasks throughout the paper, prompts follow the format of simple ${\text{noun}}$ that describes the object in the image added to ${\text{preposition}, \text{location}}$. Here, we experiment with prompts where the noun describing the object is omitted and evaluate their effectiveness in undermining the adversary’s background inpainting task. Specifically, we assumed the adversary might adjust the prompt to exclude the object (e.g., using "rocky slope" instead of "A monkey on a rocky slope") to mitigate artifacts. In all cases, \sys outperformed all baselines, as demonstrated in the Table~\ref{table:R1Q3Q4} (b). Qualitative results are depicted in Figure~\ref{fig:R1Q4}.

\begin{table}
\huge
\centering
\resizebox{\textwidth}{!}{
\begin{tabular}{c||ccc|ccc|ccc|ccc|c}
\toprule
   & \multicolumn{6}{c|}{\textbf{Foreground Inpainting}} & \multicolumn{6}{c|}{\textbf{Background Inpainting}} \\
\cmidrule(lr){2-7} \cmidrule(lr){8-13}
 \textbf{IMPRESS} & \multicolumn{3}{c|}{$\boldsymbol{m^{seg}}$} & \multicolumn{3}{c|}{$\boldsymbol{m^{bb}}$} & \multicolumn{3}{c|}{$\boldsymbol{m^{seg}}$} & \multicolumn{3}{c|}{$\boldsymbol{m^{bb}}$} \\
  & FID $\uparrow$ & Prec $\downarrow$   & LPIPS $\uparrow$  & FID $\uparrow$  & Prec $\downarrow$  & LPIPS $\uparrow$  & FID $\uparrow$  & Prec $\downarrow$   & LPIPS $\uparrow$  & FID $\uparrow$  & Prec $\downarrow$  & LPIPS $\uparrow$ & PSNR \\
\midrule

Photugard & 182.62 & 0.6510 & 0.5564 & 151.05 & 0.7990 & 0.5522 & 106.54 & 0.4954 & 0.4333 & 118.30 & 0.2158 & 0.5361 & 28.5925 \\
AdvDM & 209.21 & 0.3764 & 0.5387 & 165.99 & 0.5708 & 0.5336 & 84.63 & 0.6132 & 0.3351 & 103.76 & 0.2734 & 0.4429 & 29.1283 \\
SDST & 199.28 & 0.6252 & 0.5307 & 164.13 & 0.7432 & 0.5271 & 104.75 & 0.4852 & 0.4124 & 121.10 & 0.2130 & 0.5090 & 28.8105 \\
\midrule
\textbf{\sys} & \textbf{299.07} & \textbf{0.1614} & \textbf{0.6667} & \textbf{237.05} & \textbf{0.3300} & \textbf{0.6623} & \textbf{161.24} & \textbf{0.3230} & \textbf{0.4730} & \textbf{214.38} & \textbf{0.1360} & \textbf{0.5756} & \textbf{28.6303} \\

\bottomrule
\end{tabular}
}
\caption{Quantitative evaluation of inpainting results after applying IMPRESS~\citep{IMPRESS}.}
\label{table:IMPRESS}
\end{table}

\begin{table}
\huge
\centering
\resizebox{\textwidth}{!}{
\begin{tabular}{c||ccc|ccc|ccc|ccc|c}
\toprule
   & \multicolumn{6}{c|}{\textbf{Foreground Inpainting}} & \multicolumn{6}{c|}{\textbf{Background Inpainting}} \\
\cmidrule(lr){2-7} \cmidrule(lr){8-13}
 \textbf{Gaussian} & \multicolumn{3}{c|}{$\boldsymbol{m^{seg}}$} & \multicolumn{3}{c|}{$\boldsymbol{m^{bb}}$} & \multicolumn{3}{c|}{$\boldsymbol{m^{seg}}$} & \multicolumn{3}{c|}{$\boldsymbol{m^{bb}}$} \\
  & FID $\uparrow$ & Prec $\downarrow$   & LPIPS $\uparrow$  & FID $\uparrow$  & Prec $\downarrow$  & LPIPS $\uparrow$  & FID $\uparrow$  & Prec $\downarrow$   & LPIPS $\uparrow$  & FID $\uparrow$  & Prec $\downarrow$  & LPIPS $\uparrow$ & PSNR \\
\midrule
 Photoguard & 185.20 & 0.6808 & 0.8665 & 156.79 & 0.7814 & 0.8382 & 127.17 & 0.4322 & 0.6111 & 136.26 & 0.1958 & \textbf{0.7659} & 20.1484 \\
 AdvDM & 181.57 & 0.6730 & 0.8343 & 152.97 & 0.7864 & 0.8094 & 120.80 & 0.4460 & 0.5896 & 128.89 & 0.2084 & 0.7387 & 20.2824 \\
SDST & 185.04 & 0.6810 & 0.8507 & 154.38 & 0.7838 & 0.8228 & 123.37 & 0.4332 & 0.6006 & 135.07 & 0.2104 & 0.7546 & 20.2358 \\
\midrule
 \sys & \textbf{187.48} & \textbf{0.6682} & \textbf{0.8697} & \textbf{157.74} & \textbf{0.7804} & \textbf{0.8411} & \textbf{128.94} & \textbf{0.4056} & \textbf{0.6125} & \textbf{139.56} & \textbf{0.1820} & 0.7618 & 20.2410 \\
\bottomrule
\end{tabular}
}
\caption{Quantitative evaluation of inpainting results after applying Gaussian Noise.}
\label{table:GN}
\end{table}

\begin{table}[t!]
\huge
\centering
\resizebox{\textwidth}{!}{
\begin{tabular}{c||ccc|ccc|ccc|ccc|c}
\toprule
   & \multicolumn{6}{c|}{\textbf{Foreground Inpainting}} & \multicolumn{6}{c|}{\textbf{Background Inpainting}} \\
\cmidrule(lr){2-7} \cmidrule(lr){8-13}
 \textbf{Upscaling} & \multicolumn{3}{c|}{$\boldsymbol{m^{seg}}$} & \multicolumn{3}{c|}{$\boldsymbol{m^{bb}}$} & \multicolumn{3}{c|}{$\boldsymbol{m^{seg}}$} & \multicolumn{3}{c|}{$\boldsymbol{m^{bb}}$} \\
  & FID $\uparrow$ & Prec $\downarrow$   & LPIPS $\uparrow$  & FID $\uparrow$  & Prec $\downarrow$  & LPIPS $\uparrow$  & FID $\uparrow$  & Prec $\downarrow$   & LPIPS $\uparrow$  & FID $\uparrow$  & Prec $\downarrow$  & LPIPS $\uparrow$ & PSNR \\
\midrule
Photoguard & 136.96 & 0.8042 & 0.2476 & 111.39 & \textbf{0.8820} & 0.2562 & 60.49 & 0.8086 & 0.2639 & 62.65 & 0.5630 & 0.2842 & 30.2422 \\
AdvDM & \textbf{137.97} & 0.8078 & \textbf{0.3112} & \textbf{115.98} & 0.8844 & \textbf{0.3164} & 63.14 & 0.7886 & \textbf{0.2895} & 65.65 & 0.5428 & \textbf{0.3339} & 29.5016 \\
Mist & 136.57 & \textbf{0.8008} & 0.2474 & 112.77 & 0.8922 & 0.2576 & 61.18 & 0.7932 & 0.2632 & 64.92 & 0.5442 & 0.2823 & 30.0934 \\

\midrule
\textbf{\sys} & 137.24 & 0.8132 & 0.2784 & 115.43 & 0.8844 & 0.2851 & \textbf{65.18} & \textbf{0.7782} & 0.2840 & \textbf{66.61} & \textbf{0.5376} & 0.3068 & 29.8244 \\

\bottomrule
\end{tabular}
}
\caption{Quantitative evaluation of inpainting results after applying Upscaling method.}
\label{table:Upscaling}
\end{table}

\begin{table}[t!]
\huge
\centering
\resizebox{\textwidth}{!}{
\begin{tabular}{c||ccc|ccc|ccc|ccc|c}
\toprule
   & \multicolumn{6}{c|}{\textbf{Foreground Inpainting}} & \multicolumn{6}{c|}{\textbf{Background Inpainting}} \\
\cmidrule(lr){2-7} \cmidrule(lr){8-13}
 \textbf{JPEG} & \multicolumn{3}{c|}{$\boldsymbol{m^{seg}}$} & \multicolumn{3}{c|}{$\boldsymbol{m^{bb}}$} & \multicolumn{3}{c|}{$\boldsymbol{m^{seg}}$} & \multicolumn{3}{c|}{$\boldsymbol{m^{bb}}$} \\
  & FID $\uparrow$ & Prec $\downarrow$   & LPIPS $\uparrow$  & FID $\uparrow$  & Prec $\downarrow$  & LPIPS $\uparrow$  & FID $\uparrow$  & Prec $\downarrow$   & LPIPS $\uparrow$  & FID $\uparrow$  & Prec $\downarrow$  & LPIPS $\uparrow$ & PSNR \\
\midrule
Photugard & 178.67 & 0.7146 & 0.3830 & 144.72 & 0.8366 & 0.3790 & 101.84 & 0.5662 & 0.3736 & 117.19 & 0.2880 & 0.3969 & 29.6323 \\
AdvDM & \textbf{183.50} & \textbf{0.6800} & \textbf{0.4400} & \textbf{150.11} & 0.8126 & \textbf{0.4318} & 106.74 & 0.5394 & 0.3782 & 120.36 & \textbf{0.2614} & \textbf{0.4134} & 29.4626 \\
SDST & 179.31 & 0.7214 & 0.3956 & 145.99 & 0.8284 & 0.3914 & 104.13 & 0.5564 & 0.3783 & 118.56 & 0.2710 & 0.4003 & 29.5710 \\

\midrule
\textbf{\sys} & 183.44 & 0.6894 & 0.4126 & 149.74 & \textbf{0.8110} & 0.4080 & \textbf{108.70} & \textbf{0.5150} & \textbf{0.3837} & \textbf{124.74} & 0.2712 & 0.4084 & 29.6232 \\

\bottomrule
\end{tabular}
}
\caption{Quantitative evaluation of inpainting results after applying JPEG compression.}
\label{table:JPEG}
\end{table}

\begin{table}[t!]
\centering
\resizebox{0.22\textwidth}{!}{
\begin{tabular}{c|c}
\toprule
   & PSNR \\
\midrule
Photoguard & 31.6608 \\
AdvDM & 32.5213 \\
SDST & 32.4273 \\
\midrule
\textbf{\sys} & 32.3779 \\
\bottomrule
\end{tabular}
}
\caption{PSNR comparison of \sys and baseline methods where they are equally set with $\eta=0.06$.}
\label{table:PSNR}
\end{table}



\subsection{Robustness of \sys against purification methods}
 We conducted experiments under the same settings as outlined in the paper (~\ie~100 images, 50 prompts per image, segmentation and bounding box masks, etc.) to evaluate the robustness of \sys against the recent purification techniques, including IMPRESS~\citep{IMPRESS} and \cite{purification}. Please note that among the four suggested methods in \cite{purification}, we evaluated the two methods for which official code is available in the current time of writing this paper—Gaussian noise addition and upscaling—while the others could not be tested due to the lack of accessible implementations. For the purification methods, we follow the Pytorch implementation for JPEG compression with quality 15 and official codes for other methods where Gaussian noise strength is set to 0.05.

\begin{figure}  
    \begin{center}
    \includegraphics[width=0.95\linewidth, trim={0.5cm 7.5cm 0.5cm 7.5cm}, clip]{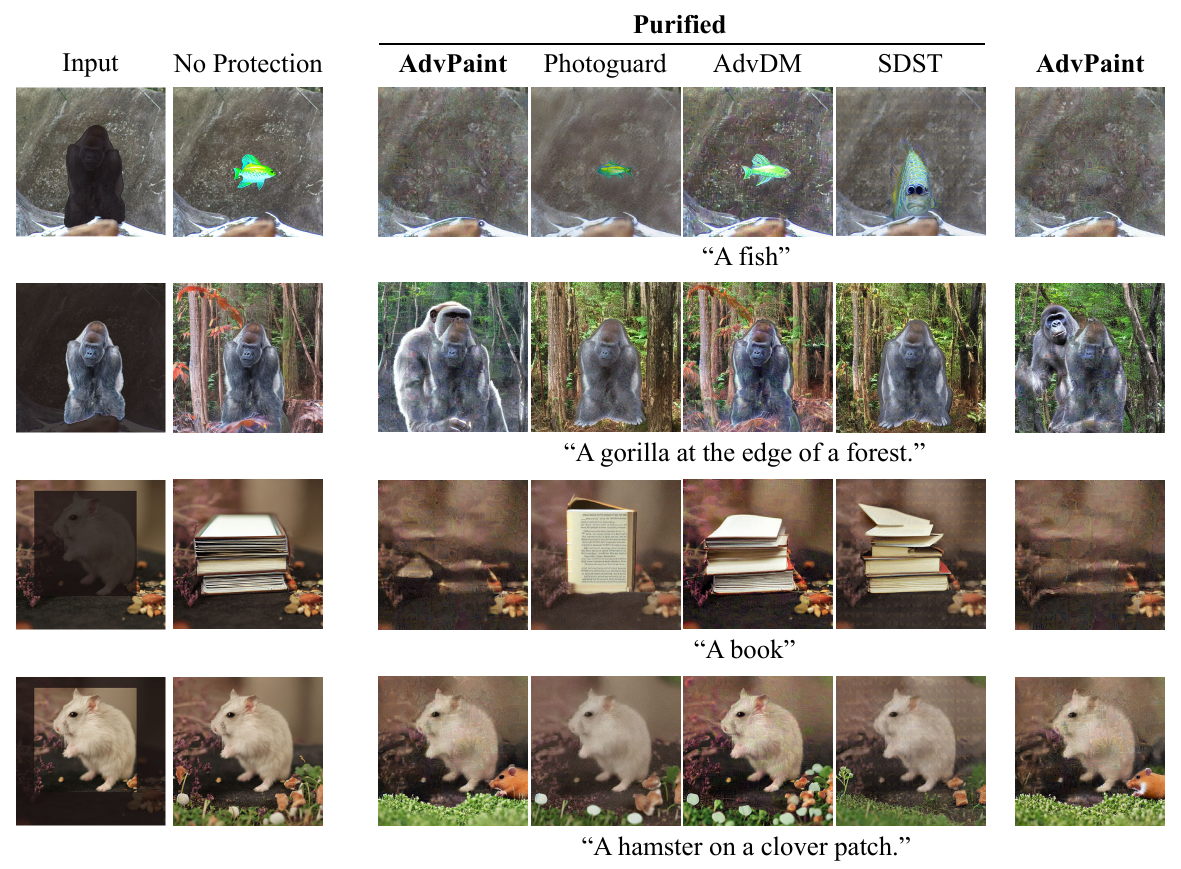}
    \end{center}
    \caption{Qualitative inpainting results after applying IMPRESS~\citep{IMPRESS}. Dark parts in the input image indicate the masked regions.}
   \label{fig:IMPRESS}
\end{figure}

 In Table~\ref{table:IMPRESS}, \sys retains its protective ability even against IMPRESS, outperforming baseline methods in terms of FID, Precision, and LPIPS. Since IMPRESS uses LPIPS loss to ensure the purified image remains visually close to the perturbed image, we believe this objective inadvertently preserves a part of the adversarial perturbation. Qualitative results are depicted in Figure~\ref{fig:IMPRESS}.

We observed that both \sys and the previous methods lose their ability to protect images when subjected to Gaussian noise addition, upscaling~\citep{purification}, and JPEG compression. In Table~\ref{table:GN},~\ref{table:Upscaling}, and~\ref{table:JPEG}, the FID, Precision, and LPIPS scores indicate significant degradation in protection under these conditions. 

\begin{figure}  
    \begin{center}
    \includegraphics[width=0.95\linewidth, trim={0.1cm 4cm 0.1cm 4cm}, clip]{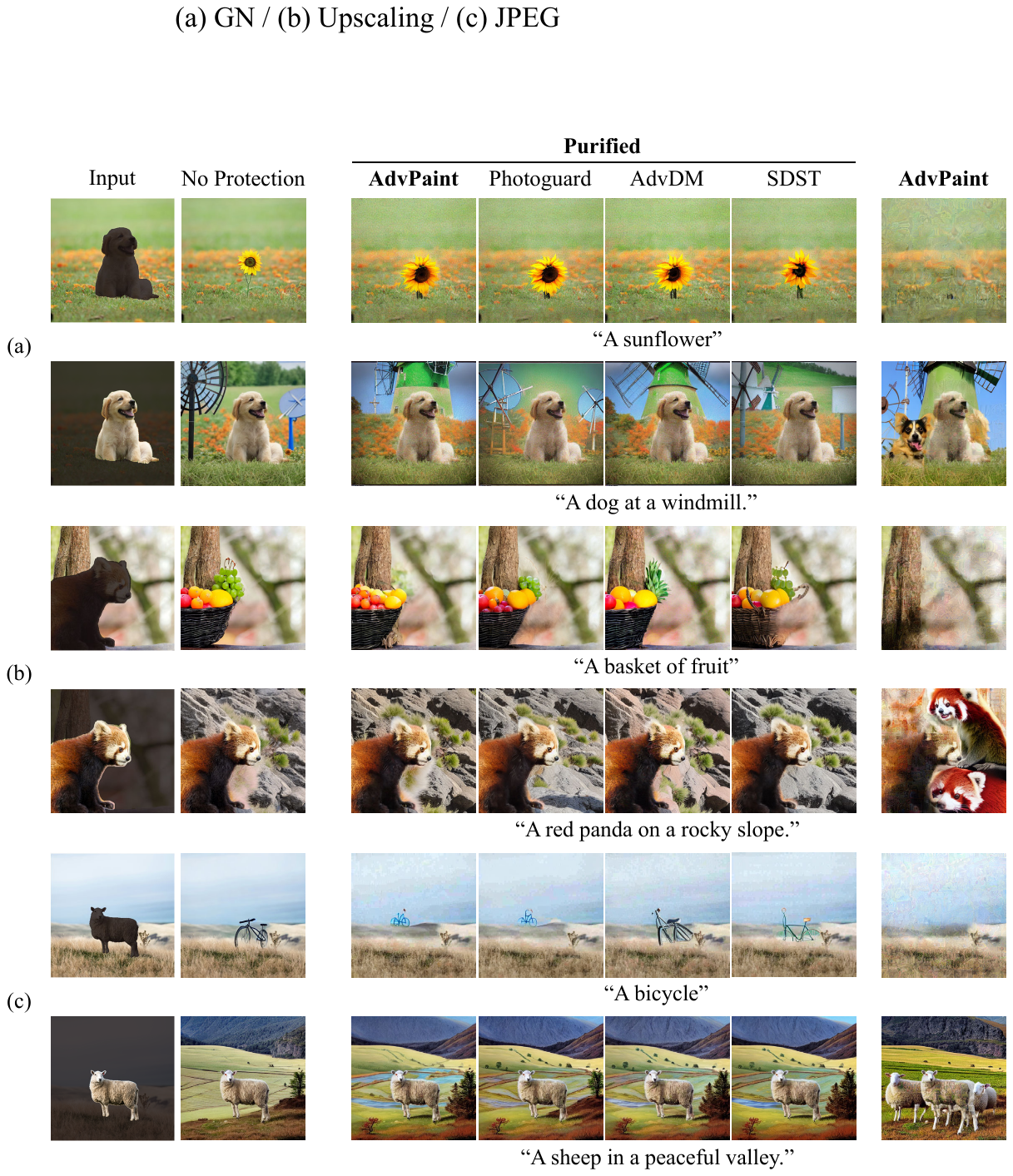}
    \end{center}
    \caption{Qualitative inpainting results after applying (a) Gaussian Noise, (b) Upscaling~\citep{purification}, and (c) JPEG compression. Dark parts in the input image indicate the masked regions.}
   \label{fig:purifications}
\end{figure}

However, as depicted in the Figure~\ref{fig:purifications} (a) and (c) regarding Gaussian noise addition and JPEG compression, \textit{the inpainted results are noisy and blurry }(e.g. noisy backgrounds for (a) “sunflower” and (c) “bicycle” images) compared to images generated from non-protected input. This raises concerns about their visual quality. It calls into question the practicality of noise-erasing methods, as the generated images often fail to meet acceptable quality standards.

Additionally, we observed a critical drawback in existing purification methods: \textit{they tend to degrade the quality of the purified image itself.} As shown in Table~\ref{table:PSNR}, \sys and baseline methods in our experiments leveraged PGD with $\eta=0.06$, ensuring adversarial examples retained a PSNR around 32 dB. On the other hand, after purification (e.g., via upscaling), we observed a PSNR drop of approximately 2.5 dB for \sys, with similar reductions observed for other methods. This decline highlights a significant trade-off between the purification effectiveness and input image quality. Qualitative results after these purification methods are depicted in Figure~\ref{fig:purifications}.

\subsection{Ablation Study of Noise Levels and Iteration Steps}

\begin{table}
\huge
\centering
\resizebox{\textwidth}{!}{
\begin{tabular}{c||ccc|ccc|ccc|ccc|c}
\toprule
   & \multicolumn{6}{c|}{\textbf{Foreground Inpainting}} & \multicolumn{6}{c|}{\textbf{Background Inpainting}} \\
\cmidrule(lr){2-7} \cmidrule(lr){8-13}
 \textbf{Noise Level $\eta$} & \multicolumn{3}{c|}{$\boldsymbol{m^{seg}}$} & \multicolumn{3}{c|}{$\boldsymbol{m^{bb}}$} & \multicolumn{3}{c|}{$\boldsymbol{m^{seg}}$} & \multicolumn{3}{c|}{$\boldsymbol{m^{bb}}$} \\
  & FID $\uparrow$ & Prec $\downarrow$   & LPIPS $\uparrow$  & FID $\uparrow$  & Prec $\downarrow$  & LPIPS $\uparrow$  & FID $\uparrow$  & Prec $\downarrow$   & LPIPS $\uparrow$  & FID $\uparrow$  & Prec $\downarrow$  & LPIPS $\uparrow$ & PSNR \\
\midrule
0.04 & 319.54 & 0.1298 & 0.6056 & 268.58 & 0.2578 & 0.6138 & 170.37 & 0.3040 & 0.4603 & 247.31 & 0.1090 & 0.5602 & 35.2832 \\
\textbf{\sys} (0.06) & \textbf{347.88} & \textbf{0.0570} & \textbf{0.6731} & \textbf{289.63} & \textbf{0.1536} & \textbf{0.6762} & \textbf{219.07} & \textbf{0.2148} & \textbf{0.5064} & \textbf{303.90} & \textbf{0.0936} & \textbf{0.6105} & \textbf{32.3779} \\
0.08 & 368.37 & 0.0320 & 0.7446 & 311.58 & 0.0992 & 0.7447 & 250.44 & 0.1630 & 0.5506 & 330.50 & 0.0782 & 0.6575 & 29.9798 \\
0.1 & 376.69 & 0.0226 & 0.7846 & 326.70 & 0.0642 & 0.7829 & 266.12 & 0.1432 & 0.5780 & 340.51 & 0.0818 & 0.6831 & 28.3171 \\
\bottomrule
\end{tabular}
}
\caption{Quantitative evaluation of inpainting results for $\eta=0.04, 0.06, 0.08, 0.1$. Results of \sys are in bolded letters.}
\label{table:eta}
\end{table}

\begin{table}
\huge
\centering
\resizebox{\textwidth}{!}{
\begin{tabular}{c||ccc|ccc|ccc|ccc}
\toprule
 & \multicolumn{6}{c|}{\textbf{Foreground Inpainting}} & \multicolumn{6}{c}{\textbf{Background Inpainting}} \\
\cmidrule(lr){2-7} \cmidrule(lr){8-13}
\textbf{Iter. Steps} & \multicolumn{3}{c|}{$\boldsymbol{m^{seg}}$} & \multicolumn{3}{c|}{$\boldsymbol{m^{bb}}$} & \multicolumn{3}{c|}{$\boldsymbol{m^{seg}}$} & \multicolumn{3}{c}{$\boldsymbol{m^{bb}}$} \\
 & FID $\uparrow$ & Prec $\downarrow$   & LPIPS $\uparrow$  & FID $\uparrow$  & Prec $\downarrow$  & LPIPS $\uparrow$  & FID $\uparrow$  & Prec $\downarrow$   & LPIPS $\uparrow$  & FID $\uparrow$  & Prec $\downarrow$  & LPIPS $\uparrow$  \\
\midrule
50 & 336.39 & 0.0826 & 0.6575 & 284.00 & 0.1872 & 0.6650 & 197.29 & 0.2736 & 0.4894 & 274.99 & 0.1122 & 0.5942 \\
100 & 343.72 & 0.0728 & 0.6720 & 287.60 & 0.1744 & 0.6781 & 207.59 & 0.2308 & 0.5087 & 296.83 & 0.0894 & 0.6082 \\
150 & 339.79 & 0.0794 & 0.6598 & 285.29 & 0.1898 & 0.6654 & 204.49 & 0.2672 & 0.4958 & 293.95 & 0.1178 & 0.5974 \\
\textbf{\sys} (250) & \textbf{347.88} & \textbf{0.0570} & \textbf{0.6731} & \textbf{289.63} & \textbf{0.1536} & \textbf{0.6762} & \textbf{219.07} & \textbf{0.2148} & \textbf{0.5064} & \textbf{303.90} & \textbf{0.0936} & \textbf{0.6105} \\
\bottomrule
\end{tabular}
}
\caption{Quantitative evaluation of inpainting results for iteration steps $=50, 100, 150, 250$. Results of \sys are in bolded letters.}
\label{table:steps}
\end{table}

\subsubsection{Analysis of Noise Levels}

In Table~\ref{table:eta}, we conducted an experiment with different values of $\eta$, ranging from 0.04 to 0.1. While the PSNR values of adversarial examples increase as $\eta$ increases, we observed consistent improvements across all evaluation metrics, including FID, precision, and LPIPS. For \sys, we set $\eta$ to 0.06, as it effectively balances protection against inpainting tasks with the quality of the protected image, achieving a PSNR of approximately 32 dB.

\subsubsection{Analysis of Iteration Steps}
In Table~\ref{table:steps}, we experimented with varying iteration steps. We evaluated iteration steps ranging from 50 to 150. Due to memory limitations, we set the default iteration steps to 250 in \sys, as higher iterations result in memory overload. The results show that while there may not be significant improvement for iteration steps around 100 and 150, optimizing for 250 steps consistently outperforms lower iteration counts, validating our choice of 250 steps as the default setting for \sys.

\subsection{Additional Quantitative Results}
\label{appendix:quan}

\begin{table}
\huge
\centering
\resizebox{\textwidth}{!}{
\begin{tabular}{c||ccc|ccc|ccc|ccc}
\toprule
 & \multicolumn{6}{c|}{\textbf{Foreground Inpainting}} & \multicolumn{6}{c}{\textbf{Background Inpainting}} \\
\cmidrule(lr){2-7} \cmidrule(lr){8-13}
 & \multicolumn{3}{c|}{$\boldsymbol{m^{seg}}$} & \multicolumn{3}{c|}{$\boldsymbol{m^{bb}}$} & \multicolumn{3}{c|}{$\boldsymbol{m^{seg}}$} & \multicolumn{3}{c}{$\boldsymbol{m^{bb}}$} \\
 Optim. Methods & FID $\uparrow$ & Prec $\downarrow$   & LPIPS $\uparrow$  & FID $\uparrow$  & Prec $\downarrow$  & LPIPS $\uparrow$  & FID $\uparrow$  & Prec $\downarrow$   & LPIPS $\uparrow$  & FID $\uparrow$  & Prec $\downarrow$  & LPIPS $\uparrow$  \\
\midrule
 LDM + $\mathcal{L}_{attn}$ & 258.63 & 0.3276 & 0.6221 & 209.34 & 0.5202 & 0.6057 & 103.89 & 0.5610 & 0.3872 & 135.07 & 0.2274 & 0.5010 \\
 
 \sys & \textbf{347.88} & \textbf{0.0570} & \textbf{0.6731} & \textbf{289.63} & \textbf{0.1536} & \textbf{0.6762} & \textbf{219.07} & \textbf{0.2148} & \textbf{0.5064} & \textbf{303.90} & \textbf{0.0936} & \textbf{0.6105} \\
\bottomrule
\end{tabular}
}
\caption{Quantitative comparison with optimization applied to the default LDM using the same objective as \sys. LDM refers to the model used in our baseline models (\eg~Photoguard, AdvDM, CAAT, etc.).}
\label{table:default LDM}
\end{table}

In Table~\ref{table:default LDM}, we conduct a simple experiment to evaluate the impact of replacing the objective functions in our baseline models. Specifically, we use the same Latent Diffusion Model (LDM) as the baselines but substitute their objective functions—replacing Eq.~\ref{eq:noise-loss} (e.g., AdvDM, CAAT) and Eq.~\ref{eq:latent-loss} (e.g., Photoguard) with our proposed attention loss (Eq.~\ref{eq:attn-loss}). Since attention blocks are also present in this \textit{default LDM}, our attention loss is directly applicable. After optimizing perturbations targeting the LDM, we generate inpainted results using the Stable Diffusion inpainting model. 

The results indicate that, while optimized with our proposed objective, the perturbations fail to provide effective protection against inpainting tasks, under-performing compared to \sys. Furthermore, compared to rows 2–6 in Table~\ref{table:inpainting_comparison} (i.e., baseline models), replacing the objective function with our attention loss does not result in a significant improvement in performance.

We attribute this to the lack of direct targeting of inpainting models, which limits their ability to counter inpainting-specific attacks. This highlights a key limitation of current protection methods that rely on the \textit{default LDM} for inpainting tasks and underscores the critical importance of designing defensive methods specifically tailored for such tasks.




\subsection{Additional Qualitative Results}
\label{appendix:qual}
\subsubsection{Single- and Multi-object images}
\begin{figure}  
    \begin{center}
    \includegraphics[width=0.95\linewidth, trim={0cm 1.4cm 0cm 1.4cm}, clip]{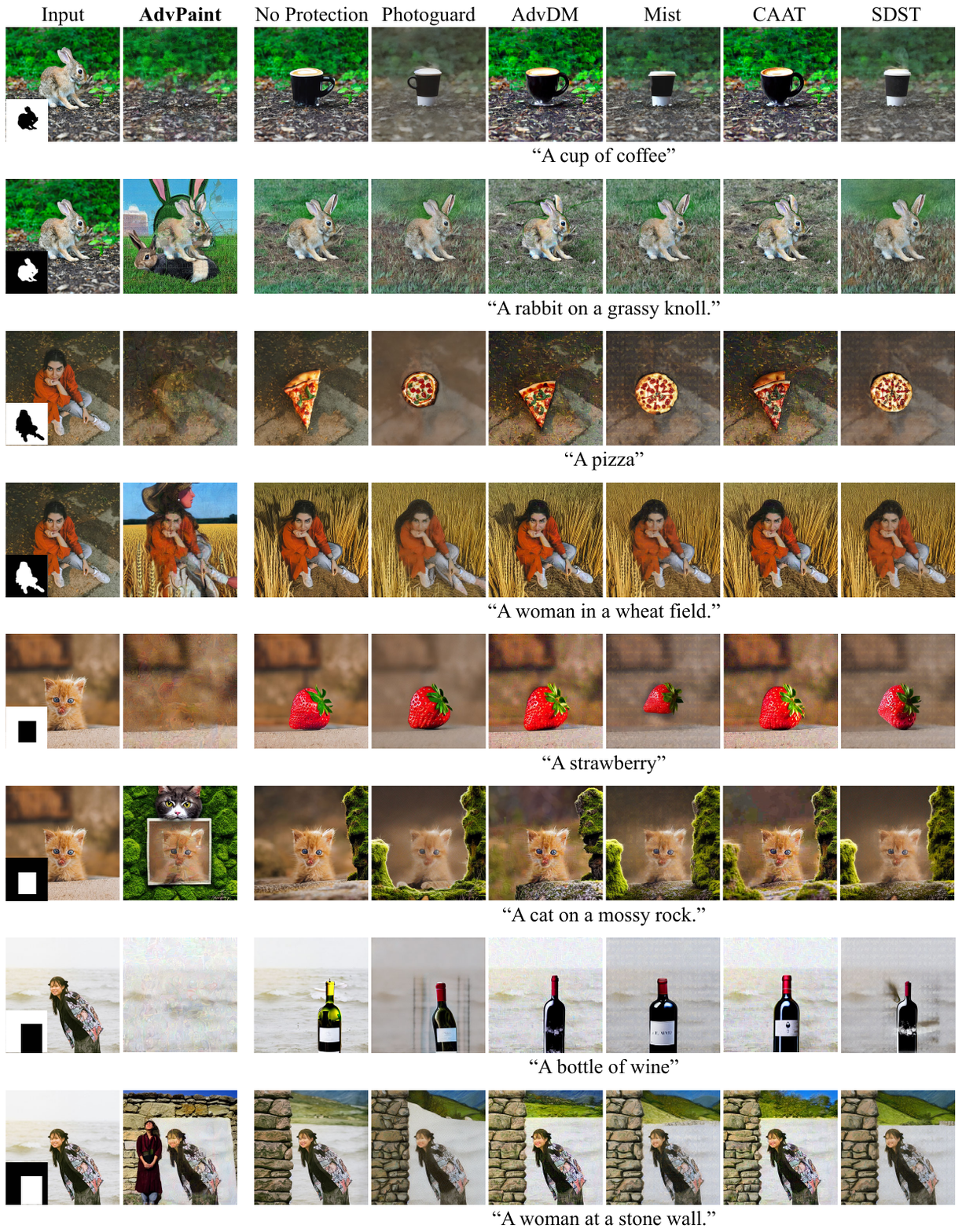}
    \end{center}
    \caption{Qualitative results of inpainting tasks using segmentation mask $m^{seg}$ and bounding box mask $m^{bb}$, comparing with prior methods.}
   \label{fig:baselines}
\end{figure}

\begin{figure}  
    \begin{center}
    \includegraphics[width=0.95\linewidth, trim={0.8cm 4.2cm 0.8cm 4.2cm}, clip]{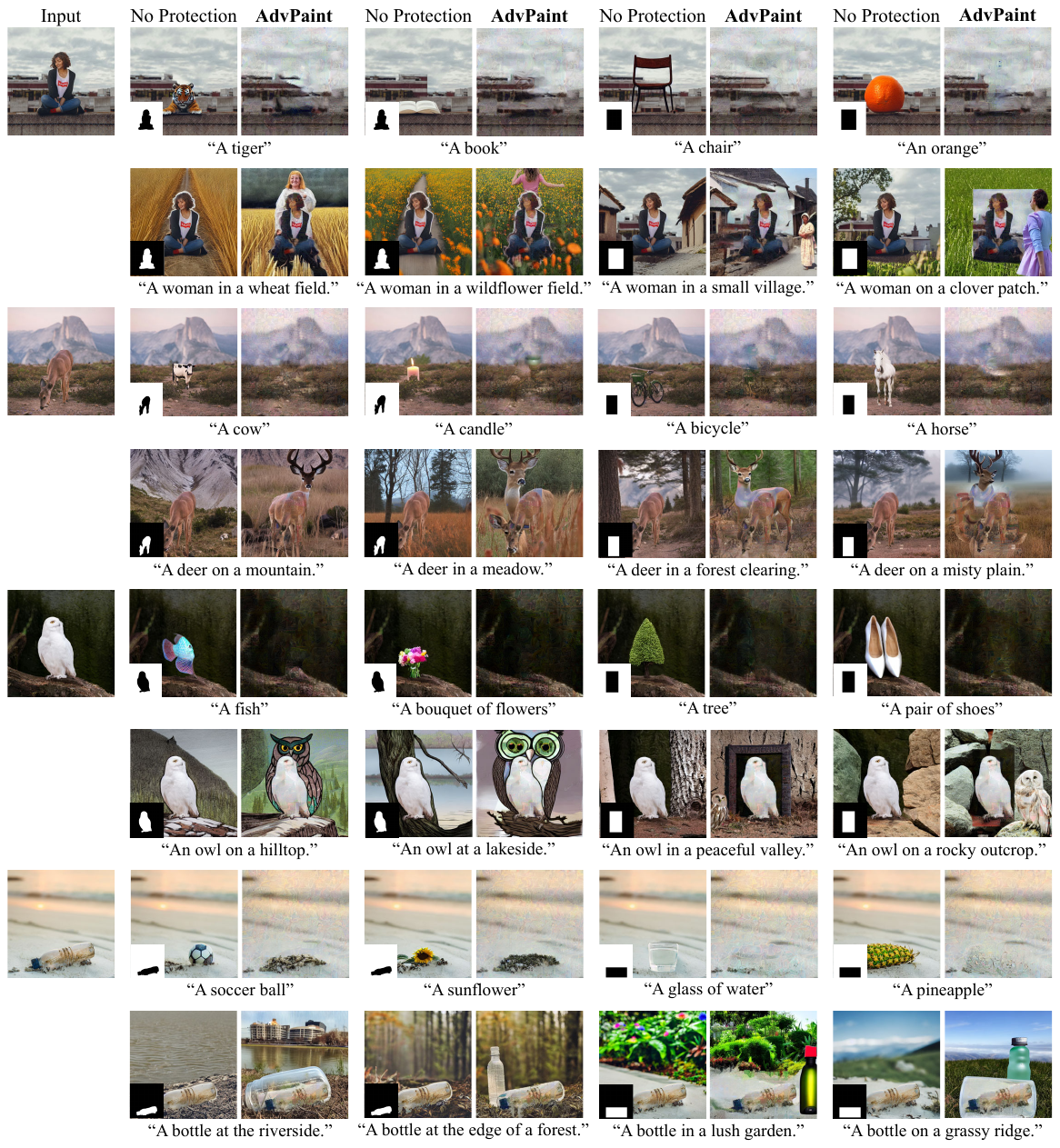}
    \end{center}
    \caption{Qualitative results of our approach on inpainting tasks with masks $m^{seg}$, $m^{bb}$, and the optimization mask $m$.}
   \label{fig:ours_appendix}
\end{figure}

We present additional qualitative results for inpainting tasks of \textit{single-object} images, comparing our method with prior protection approaches in Figure~\ref{fig:baselines}, demonstrating the effectiveness of \sys in protecting against both foreground and background inpainting with diverse masks in Figure~\ref{fig:ours_appendix}. These results confirm the robustness of our method across various masks and prompts.

\begin{figure}  
    \begin{center}
    \includegraphics[width=0.8\linewidth, trim={4cm 11.5cm 4cm 11.5cm}, clip]{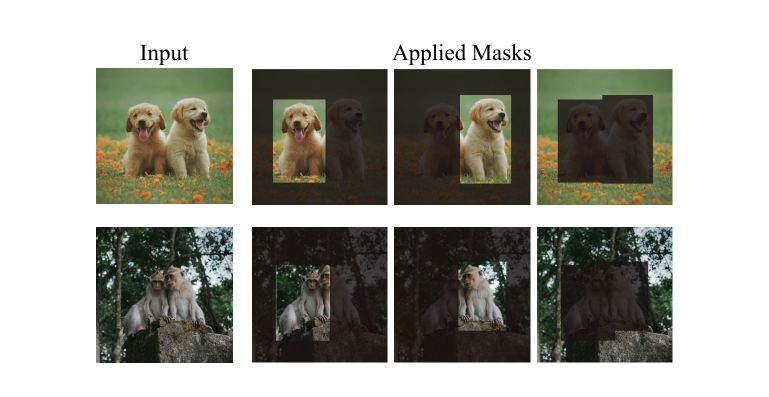}
    \end{center}
    \caption{Masks used in optimization process of multi-object images. We utilize enlarged bounding box generated by Grounded SAM. Dark parts in the input image indicate the masked regions.}
   \label{fig:multi-masks}
\end{figure}

\begin{figure}  
    \begin{center}
    \includegraphics[width=0.95\linewidth, trim={2.2cm 1.5cm 2.2cm 1.5cm}, clip]{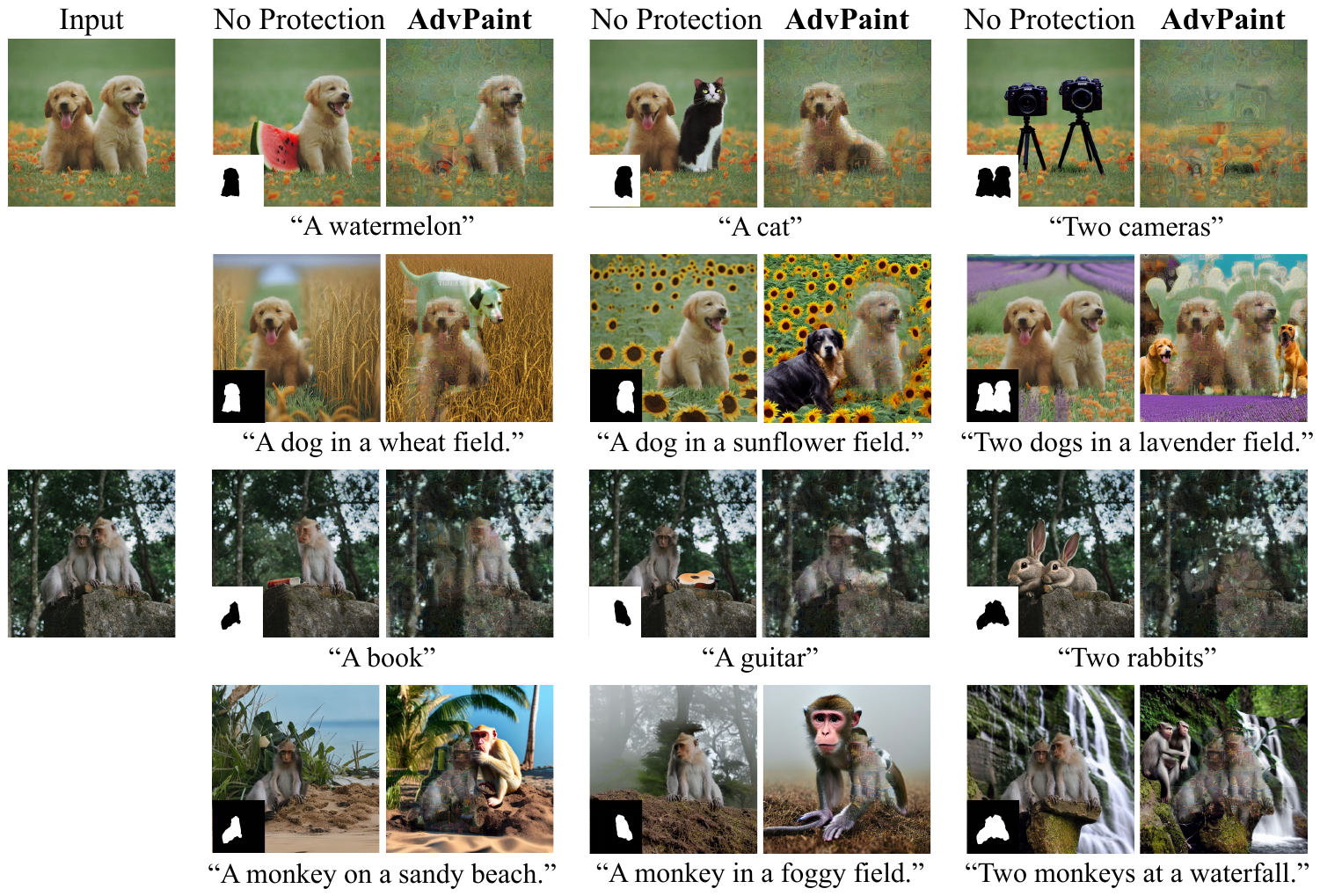}
    \end{center}
    \caption{Qualitative results of inpainting tasks for multi-object images.}
   \label{fig:multi-objects}
\end{figure}

For the optimization process of \textit{multi-object} images, we first position ourselves as content owners and select the objects that may be at risk of malicious inpainting modifications. Then, \sys performs PGD optimization for each object using enlarged bounding box masks generated by Grounded SAM. After optimizing each object, the leftover background regions, where objects potentially at risk do not exist, are also optimized. In Figure~\ref{fig:multi-masks}, we clarify the masks used to optimize multi-object images, aiding comprehension.
After securing each object, we conducted experiments with a variety of mask types, including single-object masks, masks for other objects, combined-object masks, and their inverted versions. Figure~\ref{fig:multi-objects} demonstrates the robustness of \sys for multi-object images. For example, since \sys optimizes each object individually, it ensures protection for each object, resulting in inpainted images that lack discernible objects in the foreground. Furthermore, \sys is robust to masks that encompass all objects, as shown by the absence of "two cameras" replacing "two dogs" in inpainted images. Additionally, the method effectively secures background regions when inverted masks are used for inpainting tasks. These results substantiate the effectiveness of \sys’s per-object protection method, even for complex multi-object scenarios.

\subsubsection{Alternative resources for prompt generation and mask creation}
\begin{figure}  
    \begin{center}
    \includegraphics[width=0.85\linewidth, trim={2.5cm 1.2cm 2.5cm 1.3cm}, clip]{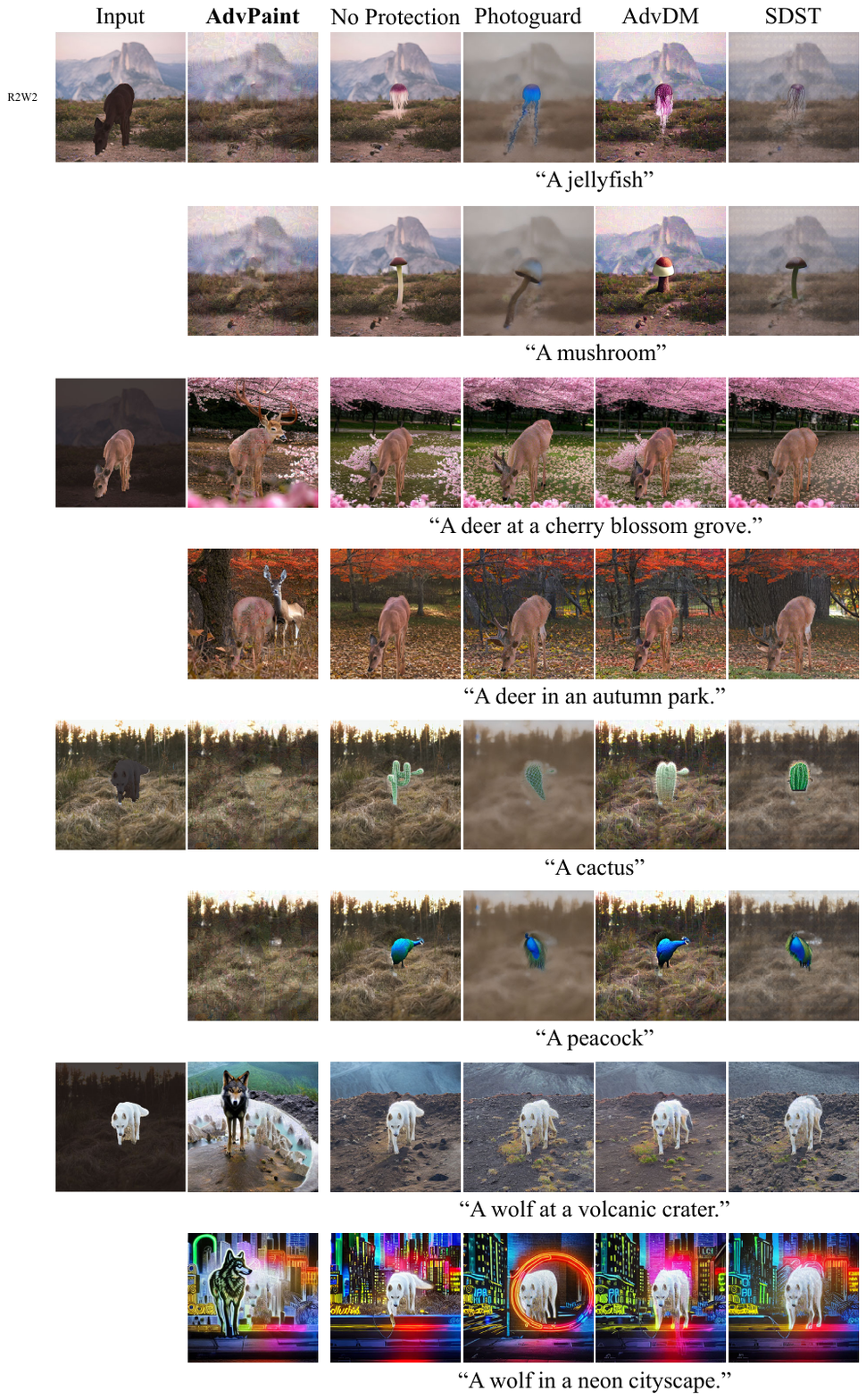}
    \end{center}
    \caption{Qualitative results of inpainting tasks with prompts and masks generated from alternative resources. Dark parts in the input image indicate the masked regions.}
   \label{fig:R2W2}
\end{figure}

\begin{figure}  
    \begin{center}
    \includegraphics[width=0.5\linewidth, trim={1.8cm 8cm 1.8cm 8cm}, clip]{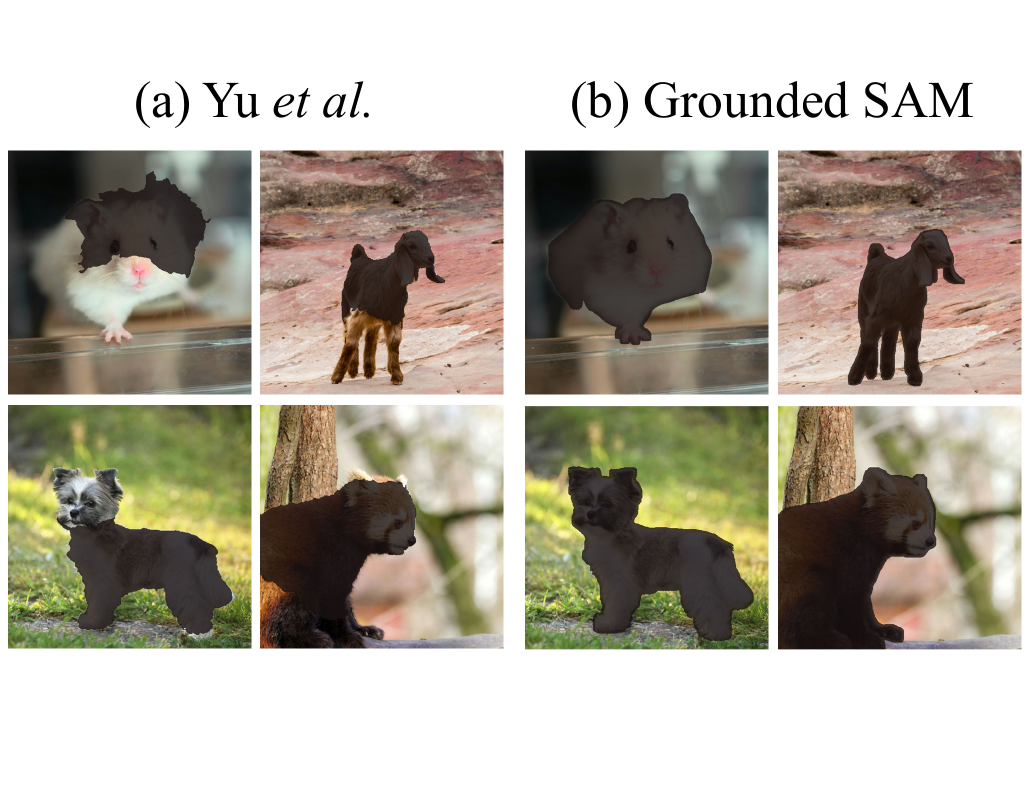}
    \end{center}
    \caption{Qualitative results of inpainting tasks with prompts and masks generated from alternative resources. Dark parts in the input image indicate the masked regions.}
   \label{fig:R2W2-weak}
\end{figure}

We conducted additional experiments employing alternative resources for prompt generation and mask creation to evaluate the robustness and generalizability of \sys’s protection performance. For prompt generation, in addition to ChatGPT, we utilized Claude 3.5 Sonnet to generate diverse prompts. For mask generation, we replaced Grounded SAM with the zero-shot segmentation method proposed by \cite{zeroshot_ris}, which employs CLIP~\citep{clip} to create object masks based on the given prompt. As depicted in Figure~\ref{fig:R2W2}, \sys retains its protection performance for inpainting tasks, comparable to its performance when using ChatGPT and Grounded SAM. However, as shown in Figure~\ref{fig:R2W2-weak}, we observe that the segmentation results from \cite{zeroshot_ris} are generally less accurate compared to those generated by Grounded SAM. This reinforces our choice of Grounded SAM as the primary segmentation tool, while also validating \sys’s adaptability to alternative segmentation approaches.

\subsubsection{Masks exceeding or overlapping the optimization boundary}
\label{appendix:real-world}
\begin{figure}  
    \begin{center}
    \includegraphics[width=0.95\linewidth, trim={2.0cm 8cm 2.0cm 8cm}, clip]{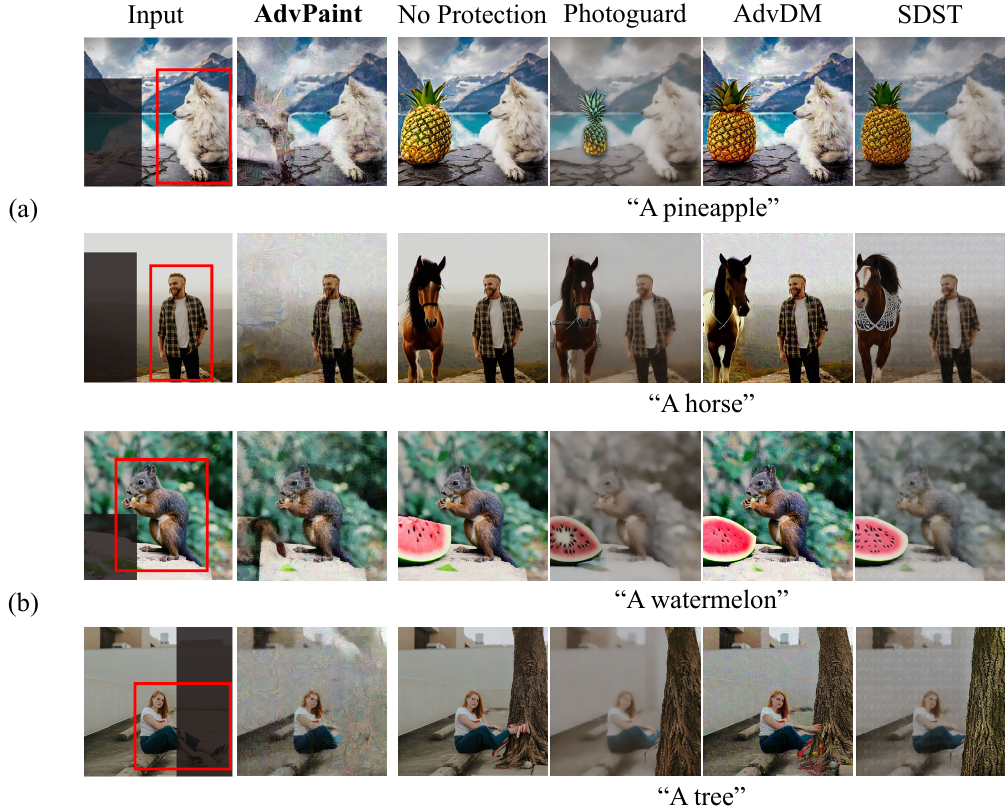}
    \end{center}
    \caption{Qualitative inpainting results where (a) masks exceed or (b) overlap with the optimization boundary (highlighted in red lines). Dark parts in the input image indicate the masked regions.}
   \label{fig:R3Q2}
\end{figure}

Since \sys leverages enlarged bounding box of the object in an image to optimize effective perturbations, one may be curious about if \sys is also robust to real-world inpainting scenarios where masks vary in sizes and shapes. In addition to the experiment conducted in Section~\ref{sec:exp_mask}, we conducted additional experiments using masks that exceed or overlap with the optimization boundary. Specifically, we visualized inpainting results where foreground masks were applied to regions without objects, simulating adversarial scenarios aimed at generating new objects in the background. As depicted in Figure~\ref{fig:R3Q2}, \sys remains robust in such diverse inpainting cases that reflect the potential threat from adversaries.

\clearpage

\end{document}